\documentclass{article}

\PassOptionsToPackage{numbers}{natbib}
  \usepackage[preprint]{neurips_2026}

\usepackage[utf8]{inputenc} 
\usepackage[T1]{fontenc}    
\usepackage{hyperref}       
\usepackage{url}            
\usepackage{booktabs}       
\usepackage{amsfonts}       
\usepackage{nicefrac}       
\usepackage{microtype}      
\usepackage{xcolor}         
\usepackage{graphicx} 
\usepackage{pifont}
\definecolor{softgreen}{RGB}{60,160,90}
\definecolor{softred}{RGB}{200,80,80}
\newcommand{\cmark}{\textcolor{softgreen}{\ding{51}}}
\newcommand{\xmark}{\textcolor{softred}{\ding{55}}}
\usepackage{caption}
\usepackage{amsmath}
\usepackage{amssymb} 
\usepackage{wrapfig}
\usepackage{subcaption}
\usepackage{tabularx}
\usepackage{float}
\usepackage{subcaption}
\usepackage{enumitem}
\usepackage[table]{xcolor}
\newcommand{\hrefgrad}[2]{%
\href{#1}{\textcolor[RGB]{65,105,225}{#2}}%
}

\title{SciReC: Diagnostic Evaluation of Multimodal, Multi-Turn Relational Reasoning\\ with Adaptive Interaction}

\author{%
  Nilay Yilmaz$^{1}$\thanks{Corresponding author: \hrefgrad{mailto:nyilmaz3@asu.edu}{nyilmaz3@asu.edu}. Code and data: \hrefgrad{https://scirecc.github.io/SciReC/}{https://scirecc.github.io/SciReC/}}\quad
  Naga Sai Abhiram Kusumba $^{2}$\thanks{Equal contribution.} \quad 
  Stella Wenxing Liu$^{1}$$^{\dagger}$\quad
  {Yezhou Yang}$^{1}$\\
   $^{1}$Arizona State University \quad 
   $^{2}$Capital One \\
}

\begin{document}

\maketitle

\begin{abstract}

Relational reasoning requires the process of perceptual understanding, comparing, and integrating the underlying relationships between concepts. This ability consists of multiple categories, such as analogical, structural, and cause-effect, each capturing a different aspect of higher-order understanding. To examine the performance of multimodal large language models (MLLM) on these relational inference tasks, we developed SciReC, a model-adaptive multimodal academic dialog benchmark. As the relational reasoning process involves multiple representations and various factors (visual understanding, exhibiting knowledge, and memory recall), we propose DMRA, a deficit-based diagnostic framework that quantifies the contribution of these components to identify the primary cause of unsuccessful cases. Claude 4.6 achieved the best performance on the overall relational score with 73\%, followed by GPT 5.4 with 68\%. Performance trends indicate that open-source models achieve their lowest scores on spatial relations, while proprietary models struggle more with hierarchical and sequential relations. Across domains, model performance is lowest on Astronomy and highest on Psychology. The results of DMRA reveal that relational reasoning is the primary source of error across all models, followed by memory limitations. 

\end{abstract}

\section{Introduction}

Relational reasoning is a fundamental factor in intelligence \cite{duncan2003intelligence}, and this capacity separates human cognition from the abilities of other animals. \cite{holyoak1995mental, oden2001ape}. This ability to extract relevant information between entities is linked to logical thinking and problem-solving in novel scenarios \cite{cattell1987intelligence, halford1998processing}. Scientific concepts in STEM education and social sciences, such as biology, physics, chemistry, sociology, and psychology, consist of examples of relational concepts that describe different levels of various scenarios \cite{goldwater2016relational}. Even basic scientific concepts require learners to develop representations of relational knowledge, and achieving expertise further depends on building interconnected concepts, classifying problems by their underlying relational structure \cite{goldwater2016relational}. The acquisition of relational concepts and their combinations, based on scientific relations, supports memory retrieval and problem-solving.

Multi-image relational reasoning in multimodal AI is evaluated through diverse tasks and datasets, including analogical reasoning, which focuses on extracting and applying relational rules, \cite{yilmaz2025voila, bitton2022vasrvisualanalogiessituation, zhang2019ravendatasetrelationalanalogical}, single-turn QA \cite{johnson2016clevrdiagnosticdatasetcompositional,nie2025mmrelbenchmarkingrelationunderstanding, wang2024muirbenchcomprehensivebenchmarkrobust}, and multi-turn dialogue \cite{liu2024mmdumultiturnmultiimagedialog, jiang2024mantisinterleavedmultiimageinstruction}, each capturing different aspects of relational structure. In a multi-turn conversation, scenarios require cross-context integration; however, many of these benchmarks rely on predefined conversation scenarios that are far from real-world dialogue interactions. In contrast, successful real-life conversations rely on the participation of both parties, so the context of the dialogue shifts dynamically in response to what each side says. However, current conversation datasets do not engage with the output of the AI models and continue the conversation with static inputs. Although they evaluate different model capabilities, their {\bf non-interactive} and {\bf one-sided} message handling often leads to incoherent and poorly grounded human-AI dialogues. To better evaluate human-AI interaction abilities along with long-context history and multimodal inputs, double-sided and model-adaptive benchmarks are necessary.

In many evaluation tasks, the atomic abilities or conditions are ignored and not included in the evaluation process, despite their influence on model performance and their potential to mislead results. This limitation is especially critical considering that all reasoning processes begin with perception \cite{peirce2012philosophical}. Visual understanding is the starting point in multimodal tasks, and without processing visual data, the models are not expected to reliably perform reasoning. This is similar to knowledge-based multimodal relational reasoning tasks, where models are required to integrate visual data with conceptual information from multiple sources. Missing information about any part of the relation prevents the model from forming a complete and accurate inference. Therefore, a diagnostic approach is essential to identify the main causes of model performance and behavior.

In response to these challenges, we introduce SciReC, a multi-image, multi-turn, model adaptive benchmark for relational reasoning in academic disciplines, which evaluates the models' performance across eight categories of relational reasoning, along with their knowledge. These relational questions require visual perception to interpret the context, memory to retain and retrieve relevant details, and the integration of knowledge to infer the underlying relationships. Our SciReC benchmark consists of the following features: (1)\textbf{ Academic context}: Instead of daily-life conversations, our benchmark focuses on university-level subject contexts, including physics, biology, and chemistry, shifting dialogue questions from general settings to knowledge-based domains. (2) \textbf{Adaptive multimodal dialogue}: Unlike prior datasets, our benchmark adjusts the conversation flow according to the models' evaluated output, which determines the context of the questions. Instead of offline conversations, it provides a dynamic structure that allows flexibility in both the number of images and the number of conversational turns. (3) \textbf{Diagnostic error analysis}: Besides evaluating the model's relational reasoning capability, we also assess the models' knowledge gap, visual understanding, and memory retention abilities. These evaluation categories support a diagnostic approach by identifying the underlying causes of failures in relational reasoning tasks. 

Our contributions and observations are summarized below:
\begin{itemize}[nosep,noitemsep,leftmargin=*]
 \item We introduce SciReC, a model-adaptive multimodal academic conversation benchmark designed to evaluate the relational reasoning of multimodal large language models across eight categories in diverse domains, incorporating three key factors: visual, knowledge-based, and memory.
 \item We evaluated state-of-the-art models on SciReC questions, achieving up to 73\% accuracy (proprietary) and 56\% (open-source). Through detailed analysis, we identify each model’s strengths and weaknesses across relational reasoning categories and scientific and social science domains.
 \item We introduce DMRA, a comprehensive diagnostic evaluation framework that identifies the root causes of failure cases through deficit-based analysis. The results indicate that relational reasoning is the primary factor across all models, followed by memory.
\end{itemize}

\section{Related Work}

\subsection{Multimodal Multi-turn Datasets}
To evaluate the multimodal dialog capability of current models, many benchmarks have been introduced. While MMDU \cite{liu2024mmdumultiturnmultiimagedialog} focuses on understanding the conversation using relations between multi-images, LoCoMo \cite{bei2026memgallerybenchmarkingmultimodallongterm}, and Mem-Gallery \cite{maharana2024evaluatinglongtermconversationalmemory} address the memory issues of the models. MMCR-Bench \cite{yan2025mmcradvancingvisuallanguage} examines the logical consistency of conversations by checking contextual referencing, and MMRC \cite{xue2025mmrclargescalebenchmarkunderstanding} evaluates memory and reasoning tasks in conversations, such as cross-turn reasoning and image management.  Mantis-Eval \cite{jiang2024mantisinterleavedmultiimageinstruction} evaluates co-reference, comparison, reasoning, and temporal understanding of models in dialogues. While MultiVerse \cite{lee2025multiversemultiturnconversationbenchmark} evaluates different key aspects of the models using a checklist-based evaluation method, ConvBench \cite{liu2024convbenchmultiturnconversationevaluation} utilizes a three-level hierarchical conversation evaluation method: perception, reasoning, and creation. Like their context, their domains also differ; some of them \cite{maharana2024evaluatinglongtermconversationalmemory, bei2026memgallerybenchmarkingmultimodallongterm, liu2024mmdumultiturnmultiimagedialog, jiang2024mantisinterleavedmultiimageinstruction, yan2025mmcradvancingvisuallanguage} consist only of real-life conversations, and others \cite{xue2025mmrclargescalebenchmarkunderstanding, lee2025multiversemultiturnconversationbenchmark, liu2024convbenchmultiturnconversationevaluation} include both scientific and daily life dialogues. Benchmarks on inter-image relations focus on daily-life contexts and lack diversity in relational types and integration of complex scientific concepts, while science-domain datasets do not evaluate relational reasoning but only the knowledge gap. SciReC bridges this gap by evaluating relational reasoning in multimodal conversations in both scientific and social science domains, as we summarized in Figure \ref{tab:datasets}.

\subsection{Evaluation Protocols and Diagnostic Error Analysis}

\begin{figure}[t]
\centering
\scalebox{0.89}{%
\begin{minipage}[t]{0.56\textwidth}
\vspace{0pt}
\centering
\scriptsize
\renewcommand{\arraystretch}{1.3}
\setlength{\tabcolsep}{1.6pt}

\begin{tabular}{l c c c c c}
\toprule
\textbf{} & \textbf{Error Analysis} & \textbf{Relation} & \textbf{Knowledge} & \textbf{Visual} & \textbf{Memory} \\
\midrule
MMCR & \xmark & \cmark & \xmark & \cmark & \xmark \\
MultiVerse & \xmark & \xmark & \cmark & \cmark & \cmark \\
Mantis-Eval & \xmark & \cmark & \xmark & \cmark & \xmark \\
MMDU & \xmark & \cmark & \xmark & \cmark & \xmark \\
Mem-Gallery & \xmark & \xmark & \xmark & \xmark & \cmark \\
MMRC & Pattern-based & \xmark & \cmark & \cmark & \cmark \\
LoCoMo & Qualitative & \xmark & \cmark & \xmark & \cmark \\
ConvBench & Hierarchical & \xmark & \cmark & \cmark & \xmark \\
\textbf{SciReC} & Deficit-based & \cmark & \cmark & \cmark & \cmark \\
\bottomrule
\end{tabular}
\end{minipage}
\hfill
\begin{minipage}[t]{0.43\textwidth}
\vspace{0pt}
\centering
\includegraphics[width=\linewidth]{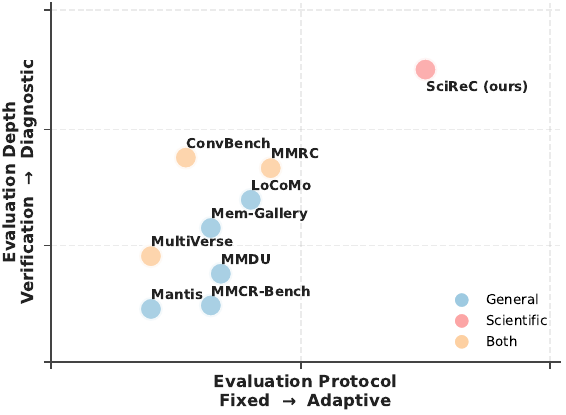}
\label{fig:compar}
\end{minipage}
}
\caption{The table compares multimodal conversational datasets in terms of context and error analysis, while the graph illustrates their comparisons across evaluation protocols and depth.}
\label{tab:datasets}
\end{figure}

Most of the multimodal multi-turn benchmarks utilized existing datasets as a data source \cite{yan2025mmcradvancingvisuallanguage, lee2025multiversemultiturnconversationbenchmark, jiang2024mantisinterleavedmultiimageinstruction, bei2026memgallerybenchmarkingmultimodallongterm, liu2024convbenchmultiturnconversationevaluation}, a few of them collected conversations from human-AI interaction \cite{xue2025mmrclargescalebenchmarkunderstanding} and AI-AI interaction \cite{maharana2024evaluatinglongtermconversationalmemory}. Although interactions make dialogs dynamic during data collection, evaluation is typically conducted in fixed conversational sequences in which the performance of the model does not influence the next turns, resulting in a static interaction flow, see Figure \ref{tab:datasets}. In contrast, SciReC provides adaptive conversational evaluation over a fixed question pool where the subsequent context is determined with a model-dependent approach. Although the question set is predefined, the evaluation process adopts a conditional interaction flow, where the selection of future question types, Knowledge Gap, Visual Perception, and Memory Retrieval, depends on whether the model’s performance exceeds a predefined threshold. By implementing this approach, SciReC enables conditionally adaptive multi-turn reasoning evaluation, rather than an offline process.

Most benchmarks in Figure \ref{tab:datasets} do not provide explicit error analysis of model performance in multimodal conversations. Their evaluation approaches primarily rely on LLM-based scoring. Although MULTIVERSE \cite{lee2025multiversemultiturnconversationbenchmark} employs a checklist-based metric for evaluation, it does not provide error analysis derived from these criteria. While LoCoMo \cite{maharana2024evaluatinglongtermconversationalmemory} identifies common errors through manual qualitative analysis, MMRC \cite{xue2025mmrclargescalebenchmarkunderstanding} conducts pattern-based analysis. ConvBench \cite{liu2024convbenchmultiturnconversationevaluation} adopts hierarchical analysis among perception, reasoning, and creation processes. Although these analyses provide detailed failure patterns, they do not attribute failures to underlying causes with systematic decomposition. Unlike these analyses, our benchmark employs deficit-based diagnostic analysis, which quantitatively identifies the relative deficit of multiple components to explore the potential error source. The method provides diagnostic insights into model failures by focusing on the weakest-performing component.

\section{SciReC Benchmark}
\label{detail_data}

\subsection{Benchmark Overview}
The SciReC benchmark is designed to evaluate the relational reasoning capabilities of current multimodal large language models in multimodal multi-turn academic conversations. This benchmark challenges models to integrate a broad range of complex concepts in two images by exploring the underlying relational structure. The datasets consist of five types of questions: relational reasoning, knowledge-based, visual understanding, memory retrieval, and memory validation, which verify whether memory-based questions are answerable using the images. We categorize relations into eight types: comparative (similarity/difference), spatial (physical location, position), sequential (temporal ordering), cause–effect (process-outcome), structural–functional (how system components function and interact), hierarchical (part–whole structures), analogical (relational mapping across domains), and other (exemplification, explanation, etc.). While relation questions demand information drawn from multiple images, other question types are specific to a single image, and figures are not provided for relational reasoning and memory questions. Figure \ref{fig:pipeline} displays the dataset creation pipeline. 

Since relational questions are extracted separately from each chapter, the number of generated dialogues equals the number of chapters. In total, 189 multimodal multi-turn conversations with 656 relational questions are generated along with ground truths extracted from the textbooks, for details please refer to Appendix \ref{dataset}. For each image, at least three knowledge-based and visual perception questions are created, along with one memory question and one memory validation question. The number of turns in each dialog varies depending on the number of relations identified in the chapters and the performance of the models, which can lead to the image questions varying across models.

\subsection{Benchmark Construction}
\paragraph{Data Extraction and Annotation}

To evaluate the relational reasoning abilities of current models across academic disciplines, we utilized 12 college-level, open-source textbooks in different domains: Biology \cite{clark2018biology2e}, Chemistry \cite{mcmurry2023organic, flowers2019chemistry2e}, Physics \cite{ling2016physics2, moebs2016physics1}, Astronomy \cite{fraknoi2022astronomy}, Economics \cite{greenlaw2022economics3e}, Psychology \cite{spielman2020psychology2e}, Behavioral Neuroscience \cite{kirby2024behavioralneuro}, and Calculus \cite{strang2016calculus, strang2016calculus2, strang2016calculus3}. The chapters and images in the textbooks are extracted, and figures are manually filtered to exclude those that do not rely on contextual knowledge. The textbook chapters serve as the source for question creation, with all necessary information about each image provided through its caption and the surrounding paragraphs. During the relation extraction and question generation processes, we employed GPT-5.4, which identified image-related content in chapters by the "Figure" tag. Relations and source texts were extracted from this textual content, \textbf{without access to the images}. Since the source texts play the key role for task evaluation, we ensure their accuracy through carefully designed prompts (see Appendix \ref{prompts}) that restrict the extraction process to only use textual content in the textbook, thereby mitigating the risk of hallucinating information not present in the textbook. In the next step, extracted relationships are validated by filtering out unclear ones by the model. Then it categorized each relation into one of eight predefined types and converted the corresponding source text into ground truths by including the relational clause linking the two figures. To avoid hallucinations, ground truths are generated to strictly follow the source text without introducing any external information, see Appendix \ref{prompts}. 

\paragraph{Question Generation.} After final validation of the results, the questions are generated utilizing the relation types, ground truths, and question templates. To prevent answer leakage and ensure structural consistency, we employed templates for both visual and relational questions. Since the knowledge-based questions obtain a clear structure and context directly extracted from the texts, we did not apply templates. A variety of questions is ensured through relational templates with several variants per relation type, and visual templates spanning eight categories with multiple question forms in each (see Appendix \ref{temp}). The model fills the placeholders (concepts, attributes, figure names) with words or phrases that appear in the ground truths to complete the question generation process. On the other hand, memory and memory validation questions are generated based on the exact facts needed to answer the relational questions. Although they share the same answers, memory validation questions are slightly paraphrased from memory questions and augmented with images to evaluate whether visual context improves model performance. \textbf{The model generates questions exclusively from the textual content, without accessing the images.} 

\begin{figure}[t!]
    \centering
    \includegraphics[width=1\linewidth]{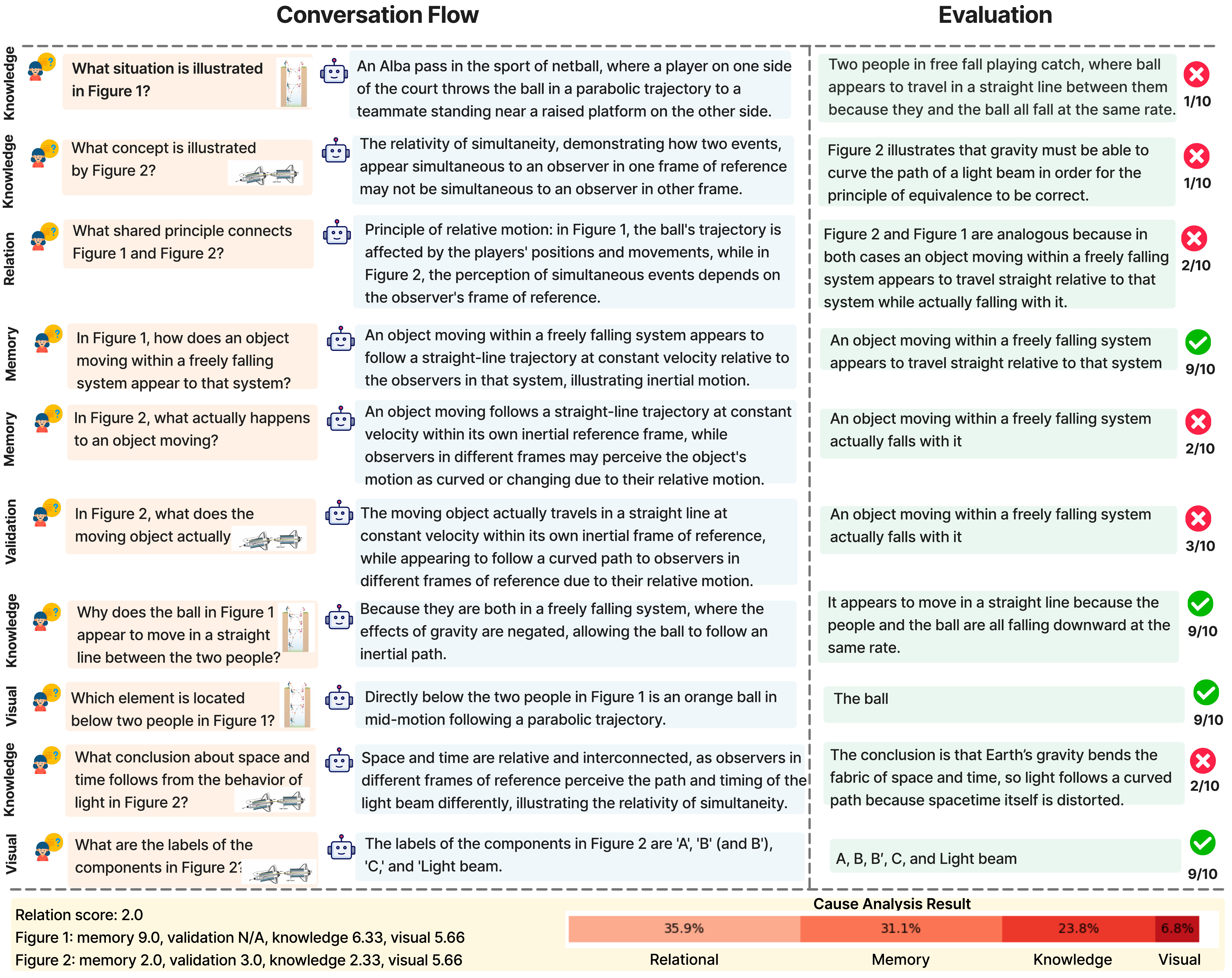}
    \caption{Example conversation flow and evaluation for an analogical relational question in the Astronomy. The panels show the question, model response, and ground truth with scores, respectively. The next question type is adjusted based on the scores. The last row shows the average score across all question types and the main failure rates. While relational reasoning is the primary cause (35.9\%), the other three factors in the bar chart are associated with Figure 2, and the remaining portion (2.4\%) corresponds to Figure 1. For better visualization of figures and detailed calculations, see Appendix \ref{calc}.}
    \label{fig:dialogue}
\end{figure}

\paragraph{Quality Control with Human Annotators.}
After verifying relations, category type, and relational questions with GPT 5.4, human annotators validated the quality of the generated relational question sets.  We selected relational Q\&A pairs: a single question from each chapter of the Biology and Physics textbooks for a multi-round manual review process by professionals. For evaluation, we defined three criteria: (1) relevance for relationships: Does the generated question accurately capture the intended relationship between the figures? (2) accuracy: Is the provided ground truth actually correct? (3) visual grounding: Can the question be answered correctly without looking at the images, relying only on generic or commonly known domain knowledge? For each round, annotators reviewed these three criteria and gave feedback about the correct and failure cases. According to these findings, we modified the prompts, dataset creation pipeline, and templates to mitigate the errors. This process ensures that the relational questions in the dataset are constructed using categorically accurate templates, grounded in reliable ground truth, and depend on visual information. Once the reviews met the quality standards, we finalized the pipeline and created the dataset. Then we \textbf{manually reviewed each relational question} alongside the corresponding figures, ground truths, and textbook source, and removed any inconsistent or unclear instances. This elimination process is applied only to relational questions, as they require integrating multiple concepts across figures, making them more error-prone, while other types can be directly and reliably extracted from textbooks.

\paragraph{Adaptive Dialogue Creation.}

The dialogue, illustrated in Figure \ref{fig:dialogue}, starts with a knowledge-based question that focuses on understanding the main concept depicted in the images based on their captions. The next turn presents a relational reasoning question between two figures without displaying the images. The model’s response is immediately evaluated using a scoring model (Claude 4.6), and subsequent question types are determined based on whether the score exceeds a predefined threshold (7/10). If the score falls below the threshold, image-specific memory retrieval questions are asked in the next turns without showing the images. If the model's performance remains below the threshold, a memory validation question is then posed for the failed image, using that image as input. Success on the validation question, despite failure on the memory question, indicates a memory retrieval issue. If both are answered incorrectly, this suggests a knowledge gap or a visual perception error. Finally, knowledge-based and visual questions are asked for both images; however, if they have already been covered in earlier turns, they are skipped, while still being included in the diagnostic analysis. The same process is repeated for the next selected relations. Since question type selection depends on model performance, the conversation becomes adaptive and model-dependent.

\subsection{Evaluation}

To evaluate the response of models, we utilized two scoring rubrics: relational and other image-specific tasks. Although the rubrics differ, they share an identical scoring scheme, with responses evaluated on a 1–10 scale: 1–3 incorrect, 4–6 partial, 7–8 mostly correct, and 9–10 complete and accurate. A score of 7 is used as the threshold for both rubrics to classify an answer as correct, with scores below 7 indicating partially correct or incorrect responses. We evaluated two models: Claude 4.6 \cite{anthropic:claude-4-6} and GPT 5.4 \cite{openai2025gpt54} for scoring the model answers. After evaluation, we selected to use Claude 4.6, which demonstrates strong alignment with human reviews. For detailed evaluation results and prompts, please refer to Appendix \ref{select} and Appendix \ref{prompts}.

\section{Deficit-Based Multimodal Relational Analysis (DMRA)}
To diagnose whether the failed responses cause from cross-image reasoning or other factors like visual encoding, knowledge gap, and memory retention, we implemented a controlled deficit-based diagnostic framework, DMRA. As the success of relational reasoning depends on the integration of these factors, we employed a two-stage error decomposition, distinguishing between upstream and relational causes. While the upstream stage assesses whether the relational reasoning is weak because of figure-level deficits, the relational stage captures cross-figure reasoning errors such as integration and abstraction. For each figure $i$, scores: memory $M_i$, knowledge $K_i$, visual $V_i$, and memory validation $MV_i$ are used to compute upstream deficits $D_x^{(i)}$ with threshold $T = 7$, (eq. \ref{eq:deficits}). 
\begin{equation}D_x^{(i)} = \max(0,\, T - X_i), \quad x \in \{M, K, V\}\label{eq:deficits}. \end{equation}
\setlength{\intextsep}{0pt}
\begin{wrapfigure}{r}{0.42\textwidth}
    \centering
    \includegraphics[width=\linewidth]{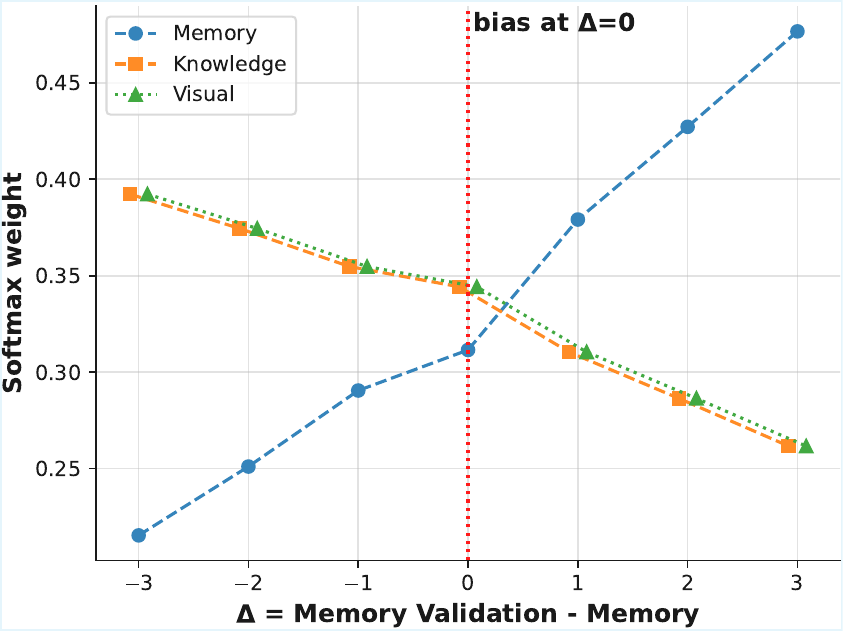}
    \caption{Softmax weights vs $\Delta$}
    \label{fig:softmax}
\end{wrapfigure}
We employ memory validation questions to isolate memory errors from perception and knowledge errors. The score difference between memory and validation questions $g_i$ dynamically adjusts the weights of each factor, see Figure \ref{fig:softmax}. We utilize the softmax function to change the weights, use equal base weights for each task $b_M = b_K = b_V = 0.33$, and select a validation effect parameter $\lambda = 0.2$. In case of the memory scores equal validation scores, we define a bias parameter $\tau = 0.1$, which slightly reduces the memory weight, as the score does not improve with images, see Equation \ref{eq:parameters}. As shown in Equation \ref{eq:soft}. The importance score of the memory task $z_M^{(i)}$ is updated based on its base weight, validation gap, and bias. Using the importance scores, the softmax function normalizes the weights of each task $w_x^{(i)}$.
\begin{equation}g_i =
\begin{cases}
MV_i - M_i, & MV_i \neq \varnothing \\
0, & MV_i = \varnothing
\end{cases} , 
\quad
\quad
\quad
\beta_i =
\begin{cases}
\tau, & MV_i \neq \varnothing \ \text{and } MV_i = M_i \\
0, & \text{otherwise}
\end{cases}\label{eq:parameters}\end{equation} 
\begin{equation}
z_M^{(i)} = b_M + \lambda g_i - \beta_i, \quad
z_K^{(i)} = b_K, \quad
z_V^{(i)} = b_V,
\quad
\quad
\quad
w_x^{(i)} = \frac{e^{z_x^{(i)}}}{e^{z_M^{(i)}} + e^{z_K^{(i)}} + e^{z_V^{(i)}}} \label{eq:soft}. \end{equation}
The weight of each figure $\alpha_i$ is calculated by normalizing each figure’s total deficit $D^{(i)}$ (with a smoothing term $\lambda_{\mathrm{fig}} = 0.5$) by the sum of all smoothed deficits of all figures $D^{(j)}$, ensuring balanced and non-zero contributions. The contribution of each upstream task $R_x^{(i)}$ is computed by multiplying the figure weight $\alpha_i$, the final task weight $w_x^{(i)}$, and task deficit score $D_x^{(i)}$ associated with figure $i$. The total upstream causes $U$ are computed by aggregating contributions of upstream tasks across all figures, see Equation \ref{eq:contribution}. 
\begin{equation}
\alpha_i = \frac{D^{(i)} +\lambda_{\mathrm{fig}}}{\sum_{j} \left(D^{(j)} + \lambda_{\mathrm{fig}}\right)}, 
\quad
\quad
R_x^{(i)} = \alpha_i\, w_x^{(i)}\, D_x^{(i)},
\quad
\quad
U = \sum_{x} \sum_i R_x^{(i)}\label{eq:contribution}. \end{equation}
For each question, the relation score $R$ is used to quantify the deviation of the answer from the threshold, where $F$ represents total failure. To find the unexplained contribution $C_{\mathrm{rel}}$ caused by relational reasoning, we subtract the explainable part from the total failure. If the upstream score is greater than or equal to the total failure, then the outcome $C_{\mathrm{rel}}$ can be attributed to limitations in the upstream task rather than to a relational error. In contrast, if the upstream score is lower than the total failure, the difference between them, $C_{\mathrm{rel}}$, corresponds to the relational error score. To share the causes proportionally within the total failure, we scale the upstream errors $s = \frac{\min(F, U)}{U}$, which prevents them from exceeding the total failure. The scaled contribution of upstream tasks $C_x^{(i)}$ is computed by multiplying their unscaled contributions by the scale parameter, refer to Equation \ref{eq:total}. 
\begin{equation} F = \max(0, T - R), \quad C_{\mathrm{rel}} = F - \min(F, U),     \quad C_x^{(i)} = s\, R_x^{(i)} \label{eq:total}. \end{equation}
The total amount of attributed failure $Total$ is calculated by summing up the upstream scores with the relational error score. To represent each task's percentage contribution to the total failure, we normalize the scores by dividing each scaled contribution of tasks to the total failure, see Equation \ref{eq:percentage}. This calculation provides a quantitative analysis of the error underlying the model's incorrect answer to the relational question. This deficit-based approach offers a diagnostic evaluation of the models' weaknesses and the primary cause of failure in the composite task.   
\begin{equation}Total = \sum_{i}\sum_{x} C_x^{(i)} + C_{\mathrm{rel}}, \quad  
C \in \{ C_x^{(i)} \} \cup \{ C_{\mathrm{rel}}\}, \quad
P = \frac{C}{Total} \times 100\label{eq:percentage}. \end{equation}

\section{Evaluation Results}
We evaluate open-sourced and proprietary models on SciReC: GPT-5.4 \cite{openai2025gpt54}, Qwen3.5 \cite{qwenteam2026qwen35omnitechnicalreport} and  Claude 4.6 \cite{anthropic:claude-4-6}, Gemma-3-27B~\citep{gemmateam2025gemma3technicalreport}, Mistral3 \cite{liu2026ministral3}, MiniCPM-V 4.5 \cite{yu2025minicpmv45cookingefficient} and InternVL3 \cite{zhu2025internvl3exploringadvancedtraining}.
\vspace{7pt}

\begin{table}[ht]
\centering
\scriptsize
\caption{Experiment results of models across relational categories. The models are ordered by their overall performance, and the color scale reflects their relative performance within each model: red indicates the lowest scores, while green indicates the highest.}
\setlength{\fboxsep}{1.4pt}
\setlength{\tabcolsep}{4pt}
\vspace{5pt}
\begin{tabular}{l |cccccccc|c}
\toprule
\textbf{Models} & \textbf{Comparative} & \textbf{Analogical} & \textbf{Sequential} & \textbf{Structural} & \textbf{Cause-Effect} & \textbf{Spatial} & \textbf{Hierarchical}& \textbf{Other} & \textbf{Overall} \\
\midrule
Claude4.6 &70.10  &80.77  &\colorbox{red!20}{ 64.52}  &\colorbox{green!20}{ 82.81 } & 79.71 &75.00  &70.00  &69.39 &73.78 \\
GPT-5.4 &65.12  &61.54  &64.52   &\colorbox{green!20}{ 78.91 }  &78.26  &66.67  & \colorbox{red!20}{ 55.00 } &59.18 & 67.99  \\
Qwen-3.5-9B &51.83 &\colorbox{green!20}{ 76.92 }  &51.61  &60.00  &66.67  & \colorbox{red!20}{ 41.67 }  &50.00  &59.18  &56.25 \\
Gemma3-27B & 45.51 &50.00  & 41.94  &\colorbox{green!20}{ 61.71 }   & 59.42 &\colorbox{red!20}{ 41.67 } &45.00  &55.10  &50.76 \\ 
Mistral3 &45.85 &57.69  &45.16  &50.78  &\colorbox{green!20}{ 57.97 }  & \colorbox{red!20}{ 33.33 }  &47.50  &51.02  &48.78 \\
MiniCPM-V &22.26 & \colorbox{green!20}{ 38.46 }  & \colorbox{red!20}{ 19.35 }  &24.80  &26.9  &33.33  &25.00  &34.69  &24.85\\
InternVL3 &15.61 &\colorbox{red!20}{ 3.85 }  & 19.35 &18.75  &21.74  &\colorbox{green!20}{ 25.00 }  &22.50  &14.29  &17.07 \\
\bottomrule
\end{tabular}
  \label{tab:results}
\end{table}
\vspace{7pt}

\subsection{Main Results}

We evaluated the relational reasoning abilities of current MLLMs on the SciReC benchmark. The Table \ref{tab:results} presents the accuracy of the models on each relational task and overall. Our evaluation shows that Claude 4.6 is the best-performing model with 73.78\% overall accuracy, achieving the highest scores for all categories. GPT-5.4 and Qwen-3.5 follow Claude4.6, yielding 68\% and 56.25\%, respectively. While Gemma3 and Mistral3 reach the closest overall performance of Qwen3-5 with 50.77\% and 48.78\% in order, other open-source models score below 24\%. The results indicate that although the gap in relational reasoning between some open-source and proprietary models is narrowing, open-source models are also separating into distinct performance tiers. For example, Qwen-3.5 performs better than GPT-5.4 on analogical relational questions and shows the same achievement for the category "Other". On the other hand, it achieved more than two times higher overall score than MiniCPM-V and InternVL3 models.

Model performance varies across relational categories, which show their weaknesses and strengths for relational reasoning. While Claude 4.6, GPT-5.4, and Gemma-3 achieve their highest score on structural relational, Claude 4.6 struggles in sequential relations (64\%), GPT-5.4 in hierarchical relations (55\%), and Gemma-3 in spatial relations (41\%). Both proprietary models excel in understanding and explaining the structure of the concept and its functions. However, GPT-5.4 has difficulty with reasoning across multiple levels of organizations, and Claude4.6 shows limitations in understanding order and step-by-step processes, same as MiniCPM-V. On the other hand, spatial relations remain challenging for many open-source models, suggesting that they lack an understanding of positional configuration within concepts. While Qwen3-5 demonstrates strong performance in analogical relations with 77\%, which transfer knowledge across different concepts, other high-performing open-source models, Gemma3 and Mistral3, achieve their highest score on cause and effect relations, which identify how changes in one factor influence another. Although analogical relations represent the highest-performing category for MiniCPM-V (38\%), they represent the lowest score for the InternVL model (3.8\%). Overall, spatial relational reasoning is a bottleneck for open-source models, whereas proprietary models show limitations in other relation types, such as sequential and hierarchical reasoning.

\begin{figure}[!t]
    \centering

    \begin{subfigure}[t]{0.50\linewidth}
        \centering
        \includegraphics[width=\linewidth]{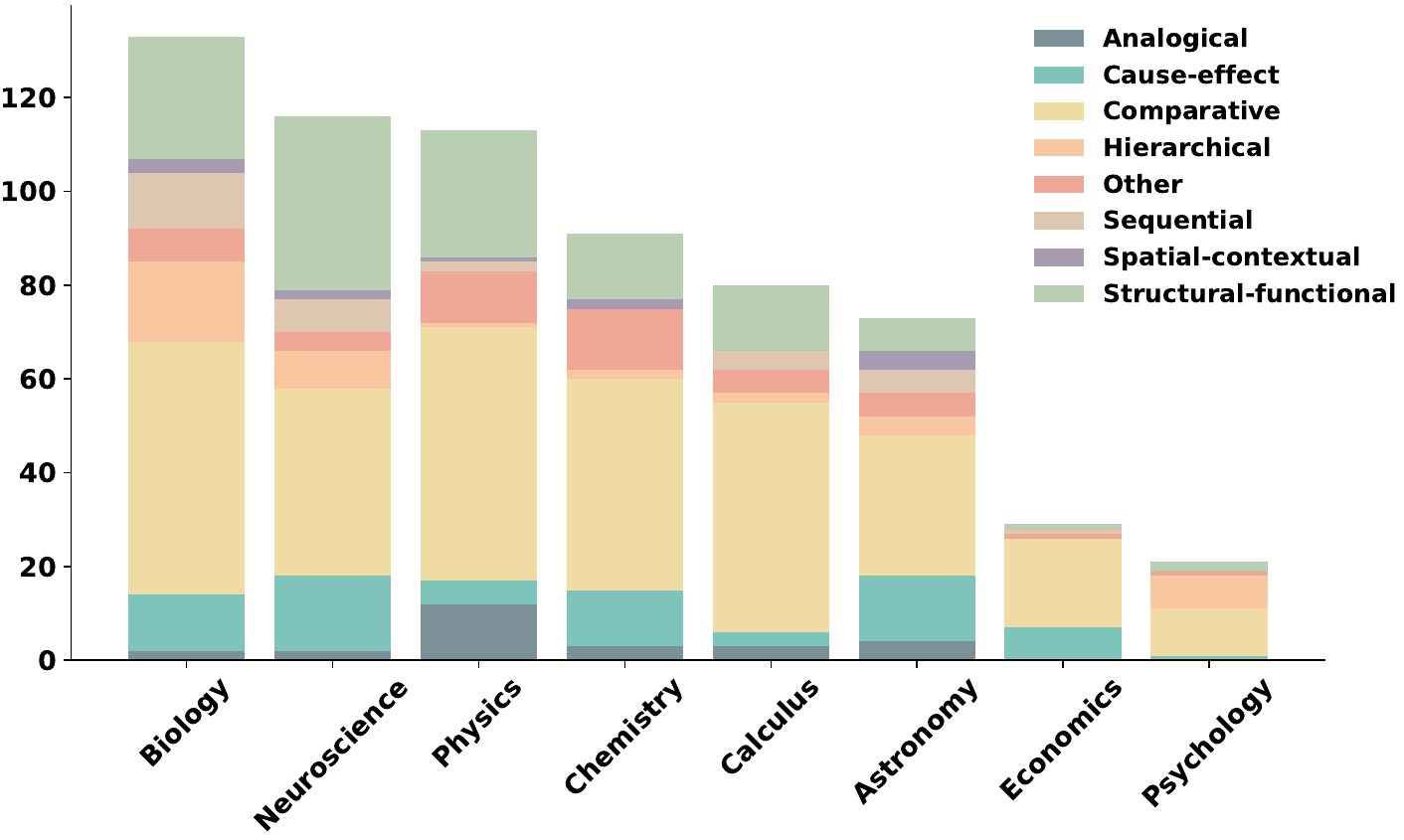}
        \caption{Distribution of relation types across domains.}
        \label{fig:textbooks}
    \end{subfigure}
    \hfill
    \begin{subfigure}[t]{0.48\linewidth}
        \centering
        \includegraphics[width=\linewidth]{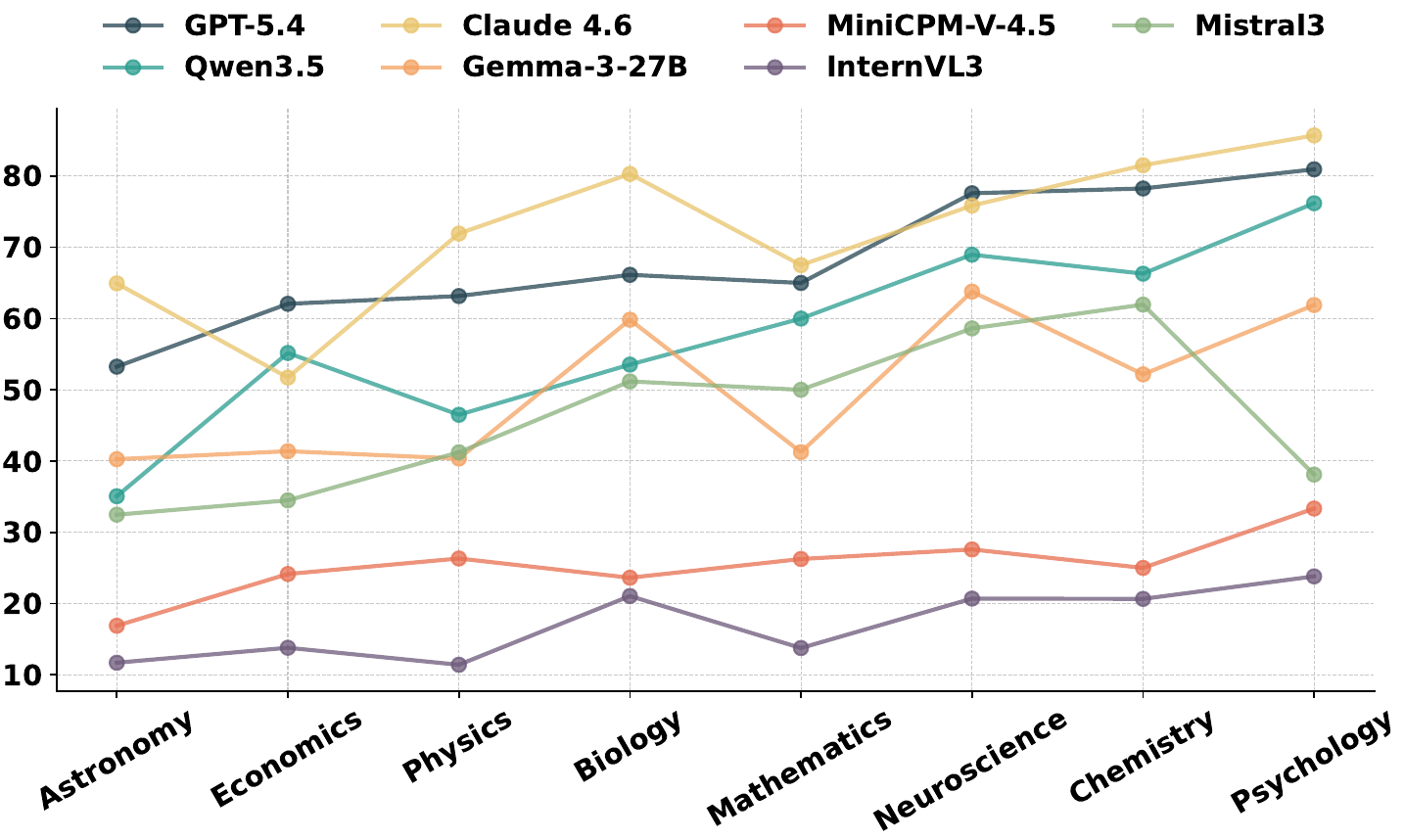}
        \caption{Performance across knowledge domains.}
        \label{fig:model_textbook}
    \end{subfigure}
    \caption{Overall comparison across domains and relation distributions.}
    \label{fig:combined}
\end{figure}

\subsection{Performance Across Domains}

SciReC consists of various relational reasoning questions across eight academic domains, see Figure \ref{fig:combined}. While domains like Biology and Behavioral Neuroscience primarily involve sequential and hierarchical relations, Physics and Chemistry do not exhibit the same inherent relational categories. Comparative relations dominate across all domains, while structural–functional and other (exemplifying, explanatory, etc.) relations are present in most domains but appear less frequently in Economics and Psychology in the dataset. Analogical relations in SciReC are mostly present in Physics and Astronomy, and spatial relations are predominantly derived from Astronomy. Calculus and Psychology questions are not mainly focused on cause-and-effect relations in SciReC. Examples of relational questions in each domain are provided in Appendix \ref{examp}.

As illustrated in Figure \ref{fig:combined}, most of the models struggle to correctly answer relational reasoning questions in the Astronomy concepts, except Claude 4.6, which shows weaker performance in Economics (52\%), where relations are mostly represented in graphs. In contrast, the Psychology domain obtains the highest accuracy for most of the models, except Mistral3, which shows the lowest performance (38\%) after InternVL3 and MiniCPM-V-4.5. Caude 4.6 obtains the highest accuracy across all domains except Economics and Behavioral Neuroscience, where GPT-5.4 outperforms it with approximately 62\% and 78\%, respectively. While Gemma-3 and InternVL3 show better performance on life sciences (Biology, Behavioral Neuroscience, and Psychology), approximately 62\% and 23\%, their performance decreases with quantitative sciences (Calculus and Physics) to 40\% and 14\%, respectively. MiniCPM-V-4.5 shows relatively consistent performance across all domains, ranging between 18\% and 33\%. While the success of Qwen3-5 in Economics, Calculus, Behavioral Neuroscience, and Psychology is close to the proprietary models up to a 10\% difference, its Astronomy, Physics, Biology, and Chemistry performances indicate a substantial gap. 

\subsection{Causal Analysis of Model Errors}

\begin{figure}[t]
    \centering
    \includegraphics[width=1\linewidth]{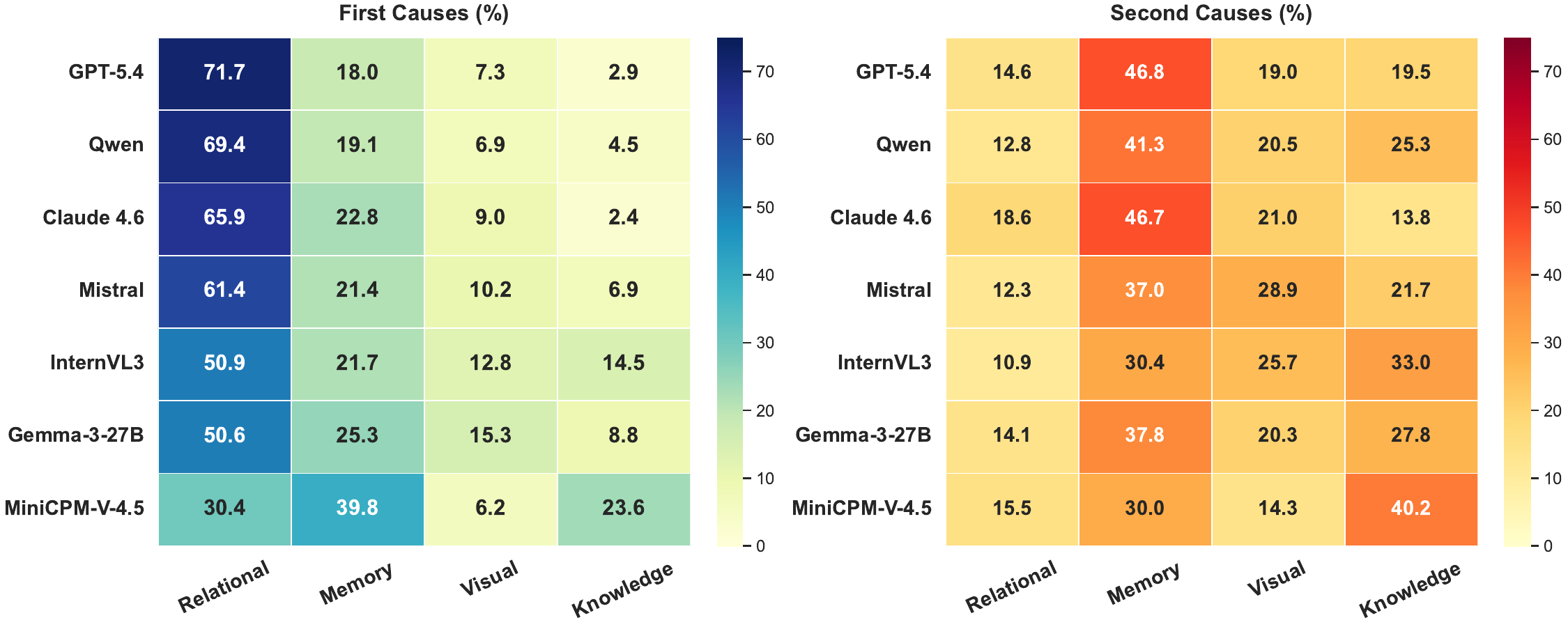}
    \caption{Results of causal analysis per model showing first and second reasons}
    \label{fig:heatmap}
\end{figure}

To analyze failures in relational questions, we use the DMRA framework with a two-stage deficit-based approach. As illustrated in Figure \ref{fig:heatmap}, the result shows that relational reasoning, which requires integration of the underlying relationship between two concepts, causes primarily performance breakdowns for all models, except MiniCPM-V-4.5. The high-performance models, GPT-5.4, Claude 4.6, Qwen-3.5, show high relational reasoning error between 66\% and 71\%, and low visual and knowledge causes below  9\% and 4.5\% respectively. This reveals that these models obtain domain-specific knowledge and understand the visual inputs; however, they struggle to connect concepts. While Mistral 3 shows similar relational reasoning (61\%) and memory-related (21\%) causality with these models, its visual and knowledge-based error rates increase to 10\% and 7\%. Relational cause drops to 51\% for InternVL3 and Gemma-3, but their knowledge gap and visual limitation increase. Memory accounts for 18–25\% as a first cause across models, except MiniCPM-v-4.5, where memory is the main challenge (39\%), followed by relational reasoning (30\%) and knowledge gap (23\%).

Across most models, memory-related issues rank as the second main cause (37–46\%); however, for InternVL3 and MiniCPM-V-4.5, knowledge gaps instead emerge as the second leading cause (up to 40\%). While knowledge gaps and limited visual understanding often emerge as secondary causes, their impact varies across models. For example, Claude 4.6 shows the lowest knowledge gap (13\%) as the second factor, whereas Qwen 3.5 exhibits a higher contribution (25\%), and GPT-5.4 shows a balanced impact (19\%) between visual and knowledge-related causes. Across other open-source models, knowledge-based deficiencies exceed visual limitations, ranging from 7\% to 26\%, except for Mistral 3, where visual limitations are higher by 7\%. Overall, the results show that strong models primarily fail at relational reasoning, whereas other models struggle more with upstream tasks.

Although strong models address the knowledge gap and visual perception errors, they struggle with memory bottlenecks, just like other models. Limitations in retrieving the relevant information from chat history and integrating it into the reasoning process act as a barrier for models to reach higher performance. These memory limitations extend to the agent-based systems that rely on MLLMs for tracking chat history, reasoning across multiple steps, and integrating past information. In agent-based settings, memory is utilized continuously across multiple interaction steps, and memory-related errors can accumulate over time, which leads to cascading failures and incorrect reasoning processes.

\section{Conclusion}
We introduce SciReC, a model-adaptive benchmark for multimodal academic dialogue, designed to assess the relational reasoning ability of MLLMs in domain-specific concepts. We include upstream tasks covering memory, knowledge, and visual reasoning, and introduce the DMRA framework, which quantitatively analyzes the main cause of failure of relational questions by a two-stage deficit-based approach. Performance trends across models indicate that structural relations achieve the highest scores among proprietary models (78-82\%), whereas sequential (64\%) and hierarchical relations(55-70\%) show the lowest. In contrast, most open source models struggle with spatial relations (33-41\%). Model performance varies across domains, with Astronomy being the most challenging and Psychology yielding the highest performance. Among incorrect responses, relational reasoning is the primary cause, ranging from 50\% to 71\%, followed by memory as the secondary cause (37–46\%). Knowledge gaps and limited visual understanding are less prominent in proprietary models but remain significant contributors to errors in most open-source models. SciReC highlights relational reasoning limitations in current models, and DMRA identifies the main causes of failures; together, they aim to contribute to improving model performance.

\paragraph{Limitations.} 
While both SciReC and DMRA provide several advantages, we identify three key limitations. (1) Our benchmark focuses on the academic domain and does not include daily life scenarios. (2) Both the question generation and model answer evaluation processes require API access keys, which may introduce additional cost. (3) The DMRA framework is designed with a three-factor error decomposition, and extending it to additional upstream factors would require reparameterization, as the softmax-based weighting is sensitive to changes in dimensionality.

\paragraph{Data License.} The data in SciReC is derived from OpenStax textbooks (CC BY-NC-SA 4.0). We provide proper attribution to the original authors and release our dataset under the same license.

\newpage

\section*{Acknowledgments}

The work was supported by ASU Enterprise Technology. We thank the Research Computing (RC) at Arizona State University (ASU) for their generous support in providing computing resources. We would like to express our gratitude to Keling Chen for providing valuable feedback that informed the development of this project. 

\bibliography{reference}
\bibliographystyle{unsrtnat}

\newpage


\appendix

\section{Dataset Details}
\label{dataset}

In total, SciReC encompasses 656 relations across eight distinct subject categories. These relations are handled in 189 conversations for these subjects. The number of turns in each dialogue is flexible, as the conversational flow depends on the model's performance. The minimum number of turns is three in the dataset, which consists of caption questions and one relational question.

\begin{wrapfigure}{r}{0.34\textwidth}
    \centering
    \includegraphics[width=\linewidth]{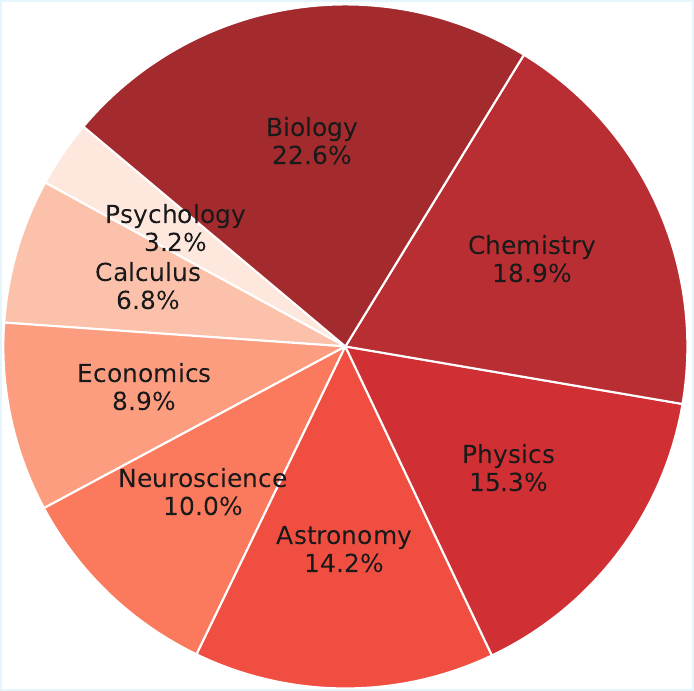}
    \caption{Distribution of dialogues}
    \label{fig:conv}
\end{wrapfigure}
The number of conversations is directly related to the number of chapters in textbooks. As relations extracted from chapters, the connections between figures and topics are preserved in one conversation. The distribution of dialogues across subjects is provided in Figure \ref{fig:conv}. The SciReC consists of 8 scientific and social science domains. While Biology covers the most conversation count in the dataset with 43 (22.6\%), the number of dialogues in Psychology is the lowest with 5 (3.2\%), followed by Calculus with 13. The combination of Organic Chemistry and Chemistry subjects encompasses 36 conversations. The count of dialogue in the Physics and Astronomy domains is similar, 29 and 26, respectively. The subjects on Behavioral Neuroscience and Economics constitute around 10\% of the total dialogues, with similar numbers, 19 and 17.

\begin{figure}[ht]
    \centering
    \includegraphics[width=1\linewidth]{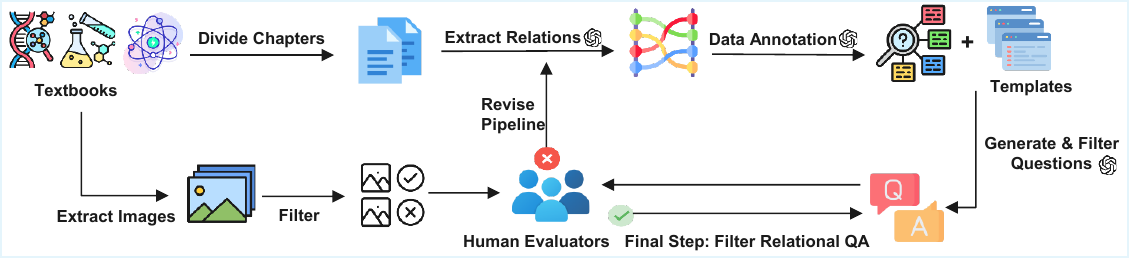}
    \caption{The benchmark creation pipeline, which accepts textbooks as input and processes them by extracting chapters and images. After extracting the relations and ground truths, questions are generated: relational and visual with templates, visual, memory, and memory validation without templates. A curated collection of filtered Q\&A pairs is evaluated in multiple rounds by human raters for three criteria (relational category match, accuracy, and visual grounding) until the pairs satisfy the established quality criteria.}
    \label{fig:pipeline}
\end{figure}

The dataset creation pipeline of the SciReC is illustrated in Figure \ref{fig:pipeline} and explained in Section \ref{detail_data}. For evaluating the open source models, we utilized an A100 GPU. The time of execution for each model takes approximately 15-18 hours, with a 4096 model context length. All data and code from SciReC are available at the following URLs:
\begin{itemize}[nosep,noitemsep,leftmargin=*]
    \item Code: \hrefgrad{https://github.com/nlylmz/SciReC}{https://github.com/nlylmz/SciReC}
    \item Data: \hrefgrad{https://huggingface.co/datasets/nlylmz/SciReC}{https://huggingface.co/datasets/nlylmz/SciReC}
\end{itemize}

\vspace{13pt}

\paragraph{Societal Impacts.} 
This paper presents a relational reasoning dataset in scientific and social science concepts, designed to assess the model's relational reasoning ability along with diagnostic causal analysis. Potential positive societal consequences include the improvement of more reliable AI tools in education, providing transparency on models' performances with reasoning failures, and providing a road map to the models based on comprehensive diagnostic results. While efficient, integration into automated pipelines like tutoring requires human oversight, as relying solely on automation may lead to misleading information.

\paragraph{LLM Usage.} 
In this study, we utilized LLMs for grammar editing, data processing and filtering, generating questions (relational, knowledge-based, visual, memory retention), evaluating baseline models' responses, running experiments with coding and debugging, understanding the technical concepts, and implementing these contexts. 

\section{Evaluation Model Selection}
\label{select}

\begin{figure}[t]
    \centering
    \includegraphics[width=1\linewidth]{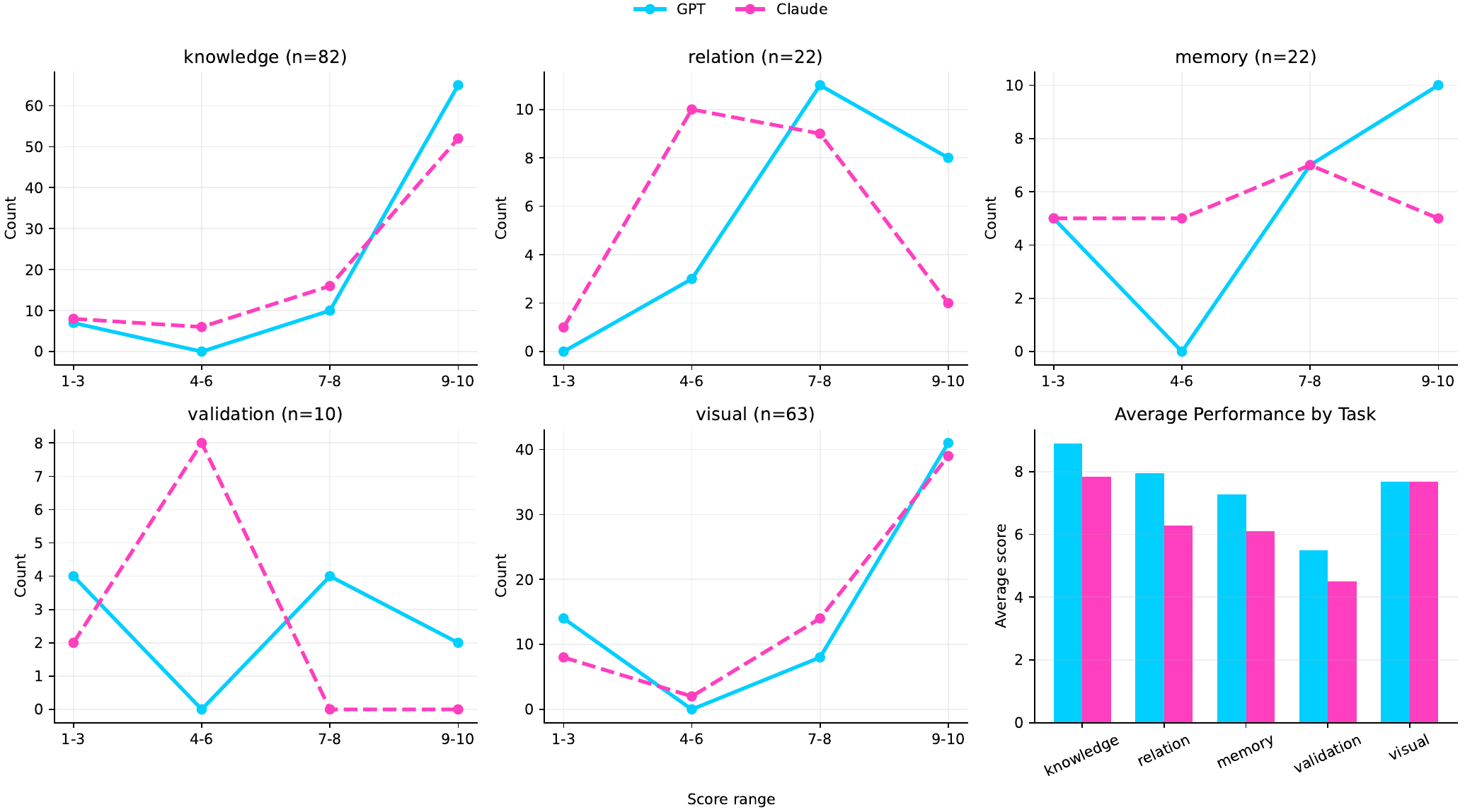}
    \caption{Distribution of evaluation scores for GPT and Claude across different task types. Line plots illustrate the number of records within each score range, and the bar chart summarizes average performance across tasks.}
    \label{fig:models}
\end{figure}

To select the scoring model aligned with the human assessment, we evaluated Claude 4.6 and GPT 5.4 with the same subset of Physics questions. By including the constraints in the prompt, we optimize the scoring for both models. As illustrated in Figure \ref{fig:models}, we evaluate the Qwen-3.5 model responses for conversational flow and score their results online with both models. The question numbers for each question types varies as the flow is conditionally dependent on the model's performance. In total, 199 model responses are evaluated, 82 knowledge-based, 63 visual, 22 relational and memory-related, and 10 memory validation. As seen in the patterns in Figure \ref{fig:models}, both models are well aligned with scoring knowledge and visual answers. For memory and validation questions, GPT-5.4 avoids using the 4-6 score range in contrast to Claude 4.6. As illustrated in relational questions, GPT-5.4 tends to give higher scores, mostly above the threshold. While the average scoring difference between models in knowledge, memory validation, and memory questions is around 1 point, for relational questions, it is 1.68. 

After observing these score differences, we selected 29 conflicting cases from the same set of questions with all types where one model scores above the threshold while the other scores below it. The aim is to select the most human-aligned scoring model for evaluating the model responses. We request human annotators to determine which score is more accurate by reviewing the question, ground truth, images, and two different scores, along with their justifications. The results show that for relational questions, GPT-5.4 provides higher scores despite the missing contextual details in the answer, although it is constrained in the prompt. Claude 4.6 not just analyzes the correctness of the relationship but also evaluates the concept integration required to explain the underlying relation between them. Based on the human review, we selected Claude 4.6 as the scoring model, which shows better alignment with human scoring.

\section{Example of DMRA Calculations}
\label{calc}

\begin{figure}[t]
    \centering
    \begin{subfigure}{0.42\linewidth}
        \includegraphics[width=0.75\linewidth]{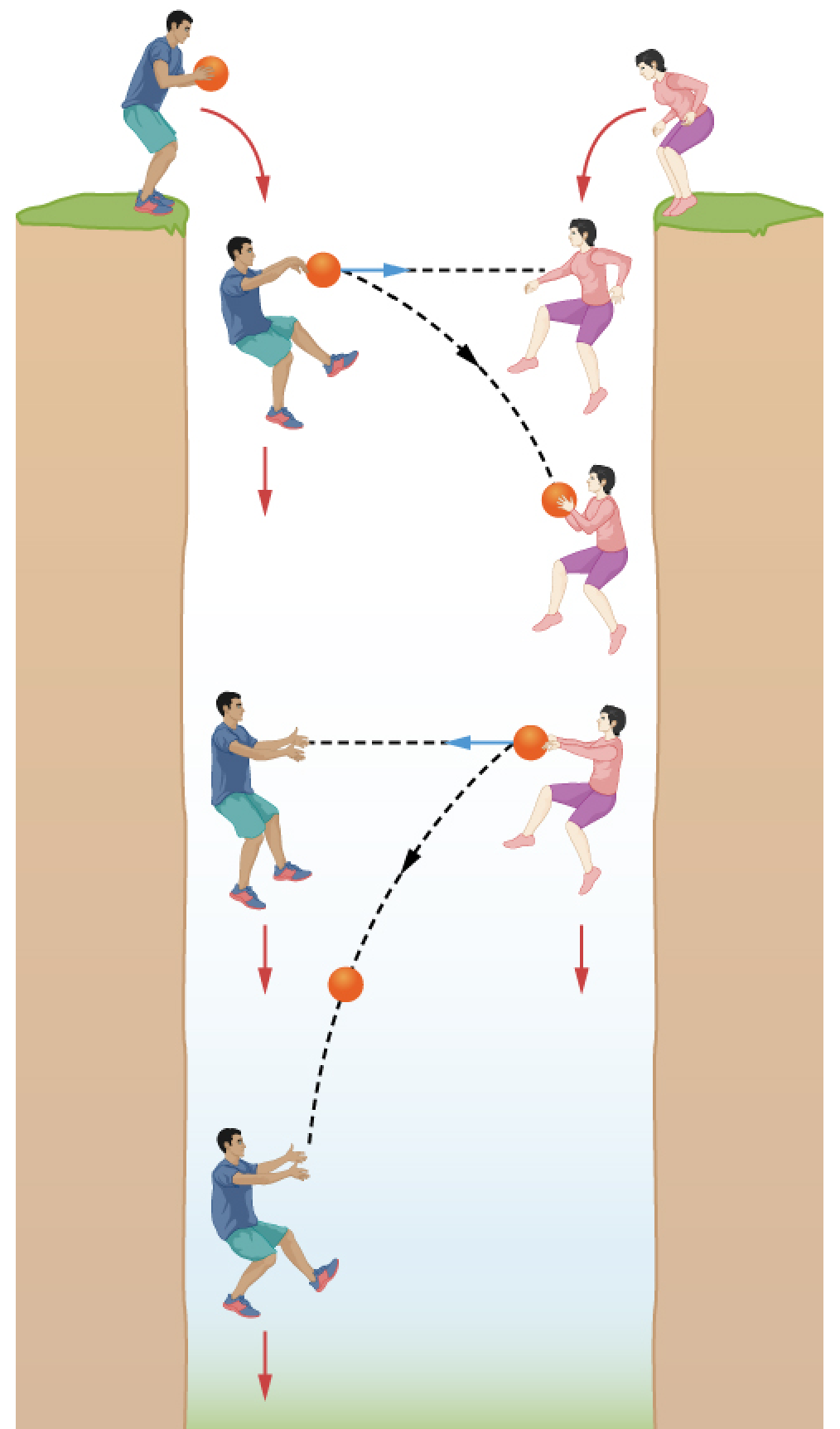}
        \caption{Figure 1}
    \end{subfigure}
    \hfill
    \begin{subfigure}{0.55\linewidth}
        \includegraphics[width=0.83\linewidth]{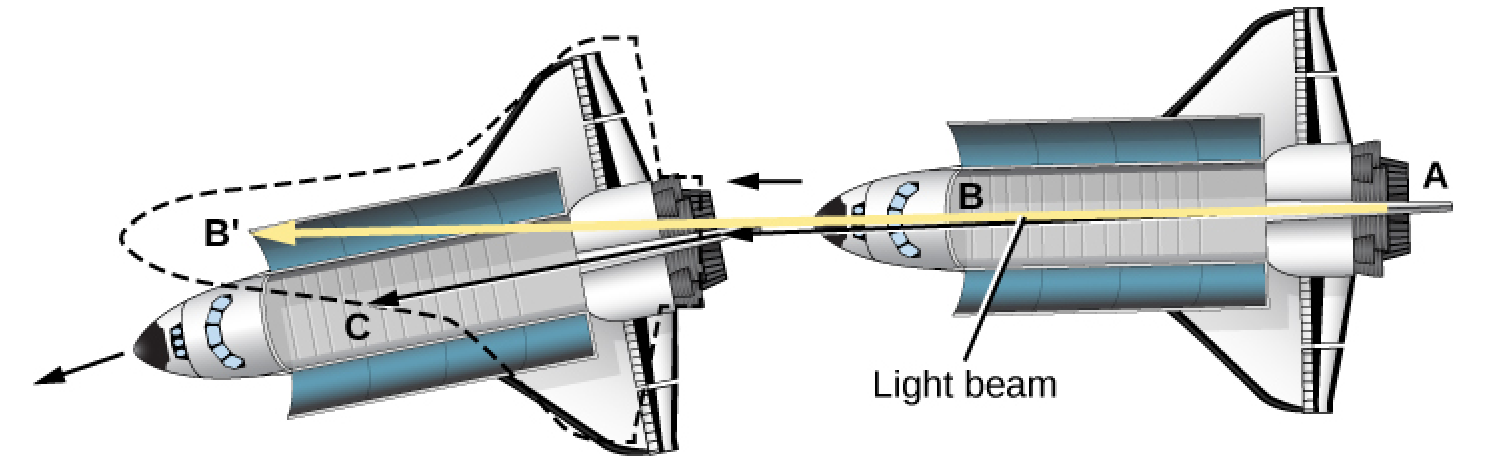}
        \caption{Figure 2}
    \end{subfigure}
    \caption{High resolution of visual images shown in Figure \ref{fig:dialogue}}
     \label{fig:main_figures}
\end{figure}

This section illustrates the steps involved in performing DMRA calculations on the given Figure \ref{fig:dialogue}. The higher resolution of figures in the relational question is provided in Figure \ref{fig:main_figures}. The illustrated part of the dialog for one relational reasoning consists of one relational, two memory-based, one memory validation, three knowledge-based, and three visual questions for each figure. The threshold for each type of question is 7, and the scores below 7 create a deficit. The score of the relational reasoning question is 2.0, and the average scores of other question types are provided in Table \ref{tab:scores_by_question_type}. As the memory score of Figure 1 is above the threshold, a memory validation question is not required.

\begin{table}[H]
\centering
\caption{Scores by question type for each figure (out of 10)}
\begin{tabular}{lcccc}
\toprule
\textbf{} & \textbf{Memory} & \textbf{Memory Validation} & \textbf{Knowledge} & \textbf{Visual} \\
\toprule
Figure 1 & 9.0 & N/A & 6.333 & 5.666 \\
Figure 2 & 2.0 & 3.0 & 2.333 & 5.666 \\
\toprule
\end{tabular}
\label{tab:scores_by_question_type}
\end{table}

The computation of the DMRA framework consists of multiple steps: 

1) As the first step, the deficit scores for each task in each figure are computed. 

Figure 1 deficits:
\begin{itemize}[nosep,noitemsep,leftmargin=*]
    \item memory: max(0, 7 - 9.0) = 0
    \item knowledge: max(0, 7 - 6.333) = 0.666
    \item visual: max(0, 7 - 5.663) = 1.333
\end{itemize}
Total deficit for Figure 1 = 2

Figure 2  deficits:
\begin{itemize}[nosep,noitemsep,leftmargin=*]
    \item memory: max(0, 7 - 2.0) = 5
    \item knowledge: max(0, 7 - 2.333) = 4.666
    \item visual: max(0, 7 - 5.666) = 1.333
\end{itemize}
Total deficit for Figure 2 = 11

2) The default weights for all tasks are 0.33. However, each task contributes differently to the result. Therefore, we adjust each task's weight for each figure based on its deficit scores using softmax. 

For Figure 1:
There is no validation gap since its memory score is above the threshold. Therefore, final weights after the softmax process are the same for all three tasks, 0.33.
\begin{itemize}[nosep,noitemsep,leftmargin=*]
    \item memory weight = 0.333
    \item knowledge weight = 0.333
    \item visual weight = 0.33
\end{itemize}

For Figure 2: 
Validation gap = 3.0 - 2.0 = 1.0 which changes memory logit to 0.33 + 0.2 × 1.0 = 0.53. The logits of knowledge and visual remain the same, 0.33, as they influence the result equally, in contrast to memory. After applying softmax to this logit, we reach the adjusted weight for each task:
\begin{itemize}[nosep,noitemsep,leftmargin=*]
    \item memory weight = 0.3792
    \item knowledge weight = 0.3104
    \item visual weight = 0.3104
\end{itemize}

3) As the deficit of each figure is not the same amount, we calculate the figure weights with $\lambda = 0.5$, which prevents ignoring the contribution of other figures. The results of the calculation show that the deficit of Figure 2 is dominant, with 95\% of the total upstream deficits.

Figure 1 weight = (2 + 0.5) / [(2 + 0.5) + (11 + 0.5)] = 0.178

Figure 2 weight = (11 + 0.5) / 14 = 0.821

4) After computing the figure and task weights, the next step is to calculate the total deficit contribution of each figure and all upstream tasks.

Figure 1 total deficits:
\begin{itemize}[nosep,noitemsep,leftmargin=*]
    \item memory: 0.178 * 0.333 * 0 = 0
    \item knowledge: 0.178 * 0.333 * 0.666 = 0.039
    \item visual: 0.178 * 0.333 * 1.333 = 0.079
\end{itemize}
Total : 0 + 0.039 + 0.079 = 0.119

Figure 2 total deficits:
\begin{itemize}[nosep,noitemsep,leftmargin=*]
    \item memory: 0.821 * 0.379 * 5 = 1.557
    \item knowledge: 0.821 * 0.31 * 4.666 = 1.19
    \item visual: 0.821 * 0.31 * 1.333 = 0.34
\end{itemize}
Total : 1.557 + 1.190 + 0.34 = 3.087

Upstream deficits:
\begin{itemize}[nosep,noitemsep,leftmargin=*]
    \item memory: 0 + 1.557 = 1.557
    \item knowledge: 0.039 + 1.19 = 1.229
    \item visual: 0.079 + 0.34 = 0.419
\end{itemize}
Total: 1.557 + 1.229 + 0.419 = 3.206

5) Total upstream deficit represents the explained error score by knowledge gap, limited visual understanding, and memory issues. To find the unexplained, rational part, we need to use the score (2.0) from the relational reasoning question. 

Total failure = 7 - 2 = 5

Upstream total = min (5 , 3.206 ) = 3.206
Relational = 5 - 3.206 = 1.793

6) As the upstream total is equal to the upstream deficits, they do not require normalization. 
\begin{itemize}[nosep,noitemsep,leftmargin=*]
    \item Figure 1 Memory = 0
    \item Figure 1 Knowledge = 0.039
    \item Figure 1 Visual = 0.079
    \item Figure 2 Memory = 1.557
    \item Figure 2 Knowledge = 1.19 
    \item Figure 2 Visual = 0.34
    \item Relational = 1.793
\end{itemize}

By dividing each deficit contribution to the total failure, we reach the final percentages of each factor.

\begin{itemize}[nosep,noitemsep,leftmargin=*]
    \item Figure 1 Memory = ( 0 / 5 ) * 100 = 0\%
    \item Figure 1 Knowledge = ( 0.039 / 5) * 100 = 0.8\%
    \item Figure 1 Visual = ( 0.079 / 5 ) * 100 = 1.6\%
    \item Figure 2 Memory = ( 1.557 / 5 ) * 100 = 31.14\%
    \item Figure 2 Knowledge = ( 1.19 /5 ) * 100 = 23.8\%
    \item Figure 2 Visual = ( 0.34 / 5 ) * 100 = 6.8\%
    \item Relational = ( 1.793 / 5 ) * 100 = 35.9\%
\end{itemize}

The results show that most of the failures were caused by relational reasoning, followed by Figure 2 memory and Figure 2 knowledge. The tasks in Figure 1 have minimal impact on the results due to their low deficits and low figure weight.

\section{Examples of Relational Questions}
\label{examp}

\begin{figure}[H]
    \centering
    \includegraphics[width=1\linewidth]{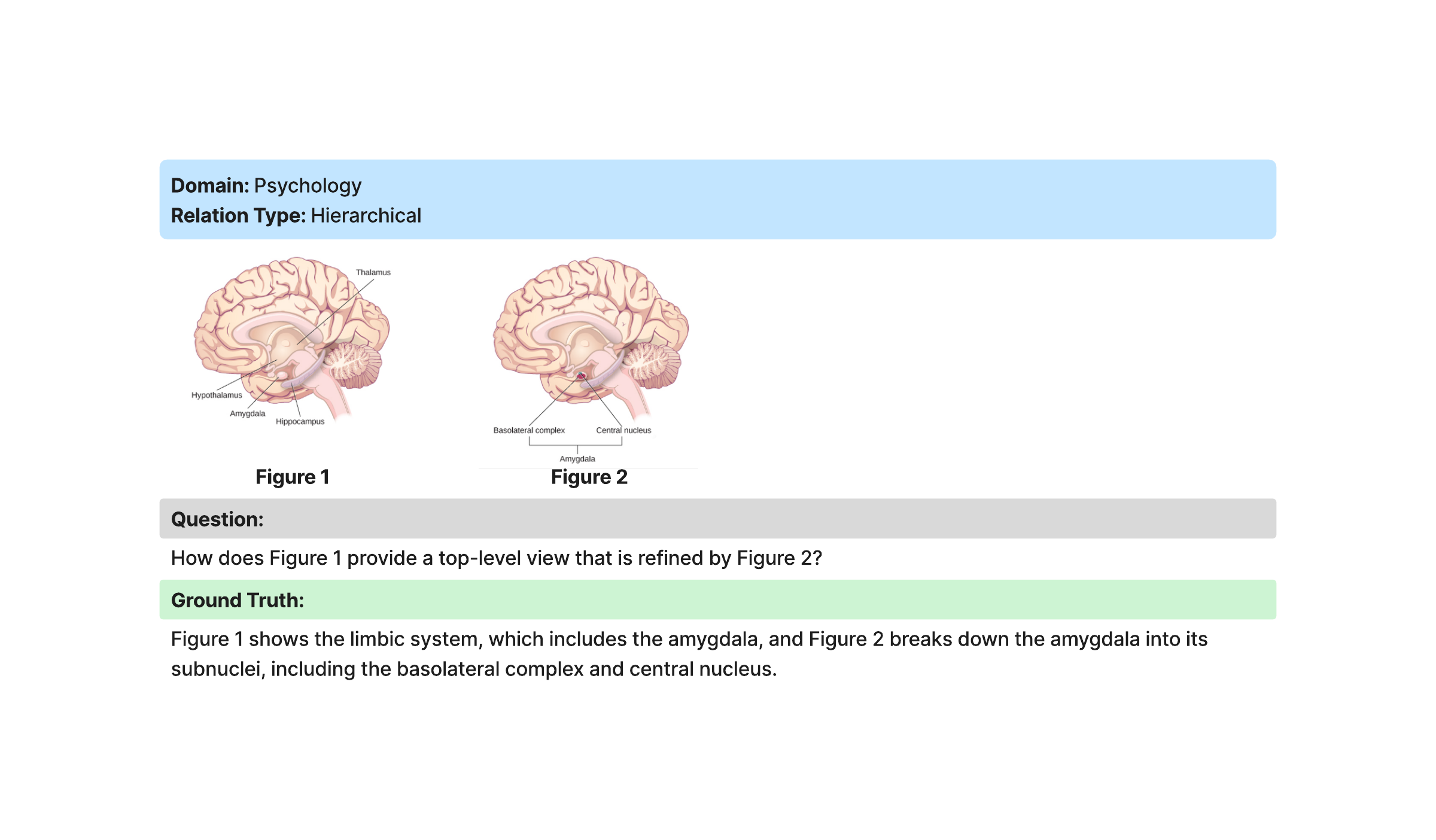}
    \caption{Example of a hierarchical relational question in the Psychology domain }
    \label{fig:placeholder1}
\end{figure}

\begin{figure} [H]
\centering
    \includegraphics[width=1\linewidth]{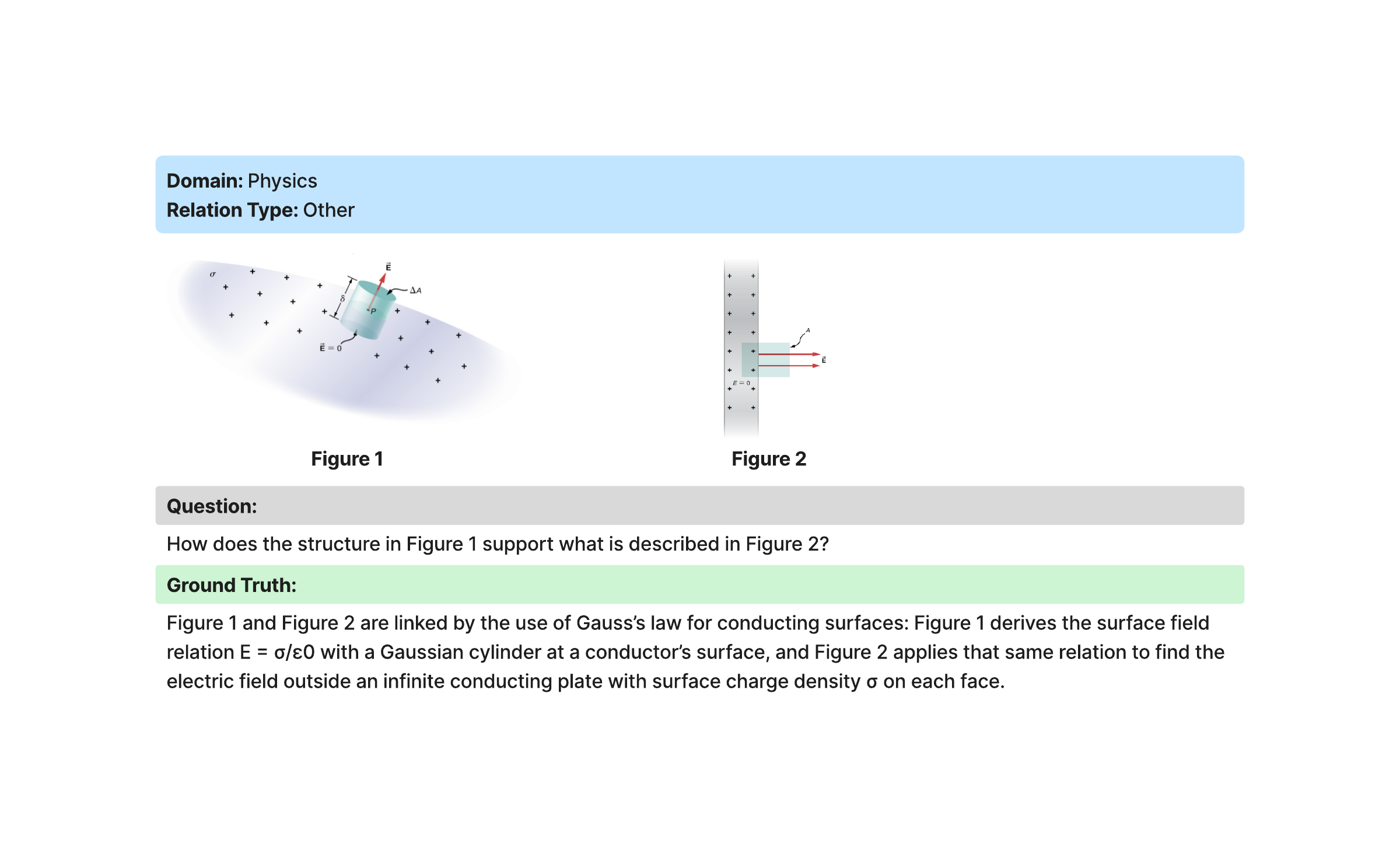}
    \caption{Example of "Other" type of relational question in the Physics domain}
    \label{fig:placeholder2}
\end{figure}

\begin{figure}[H]
    \centering
    \includegraphics[width=1\linewidth]{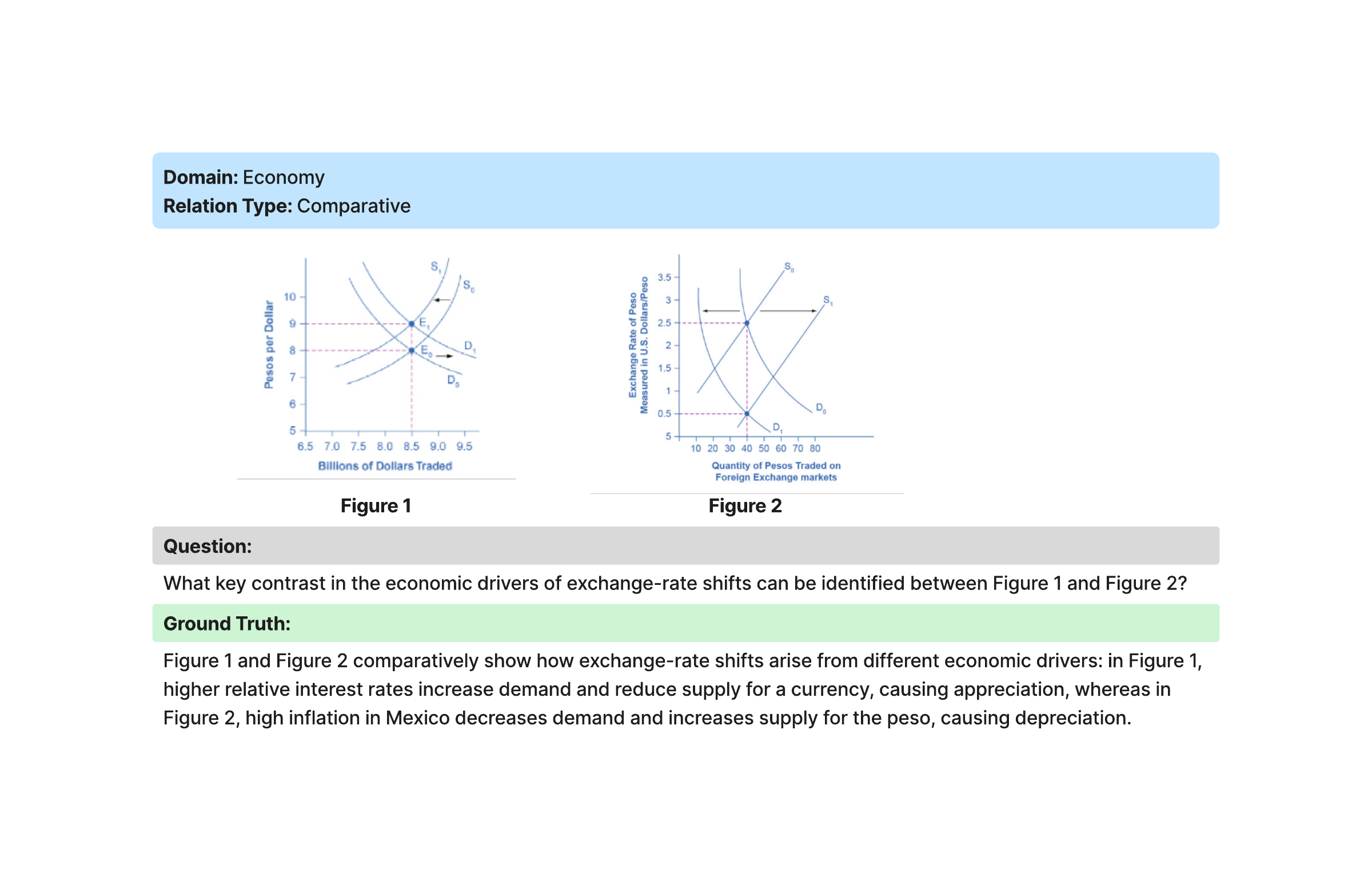}
    \caption{Example of a comparative relational question in the Economics domain}
    \label{fig:placeholder3}
\end{figure}

\begin{figure}[H]
    \centering
    \includegraphics[width=1\linewidth]{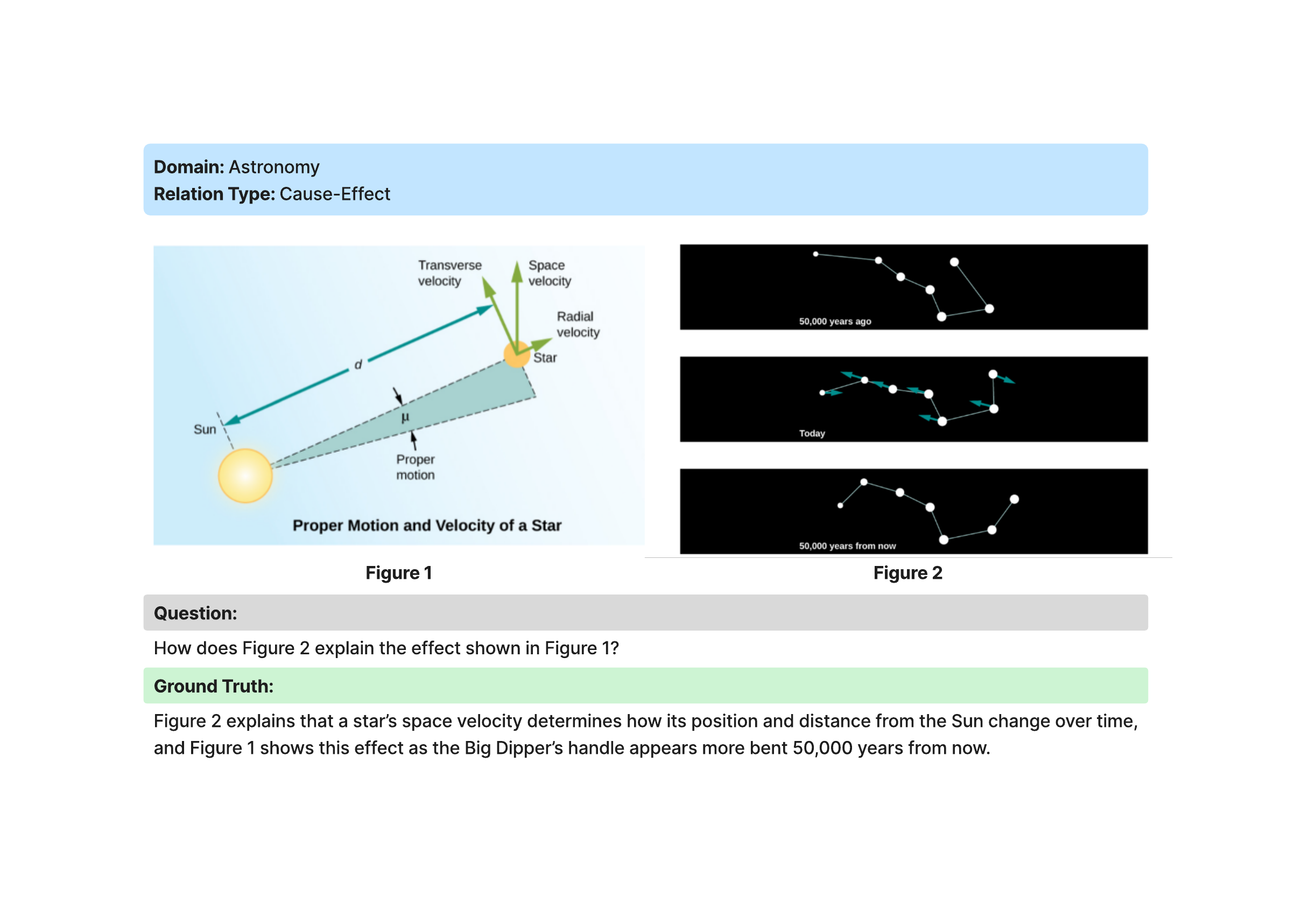}
    \caption{Example of a cause and effect relational question in the Astronomy domain}
    \label{fig:placeholder4}
\end{figure}

\begin{figure}[H]
    \centering
    \includegraphics[width=1\linewidth]{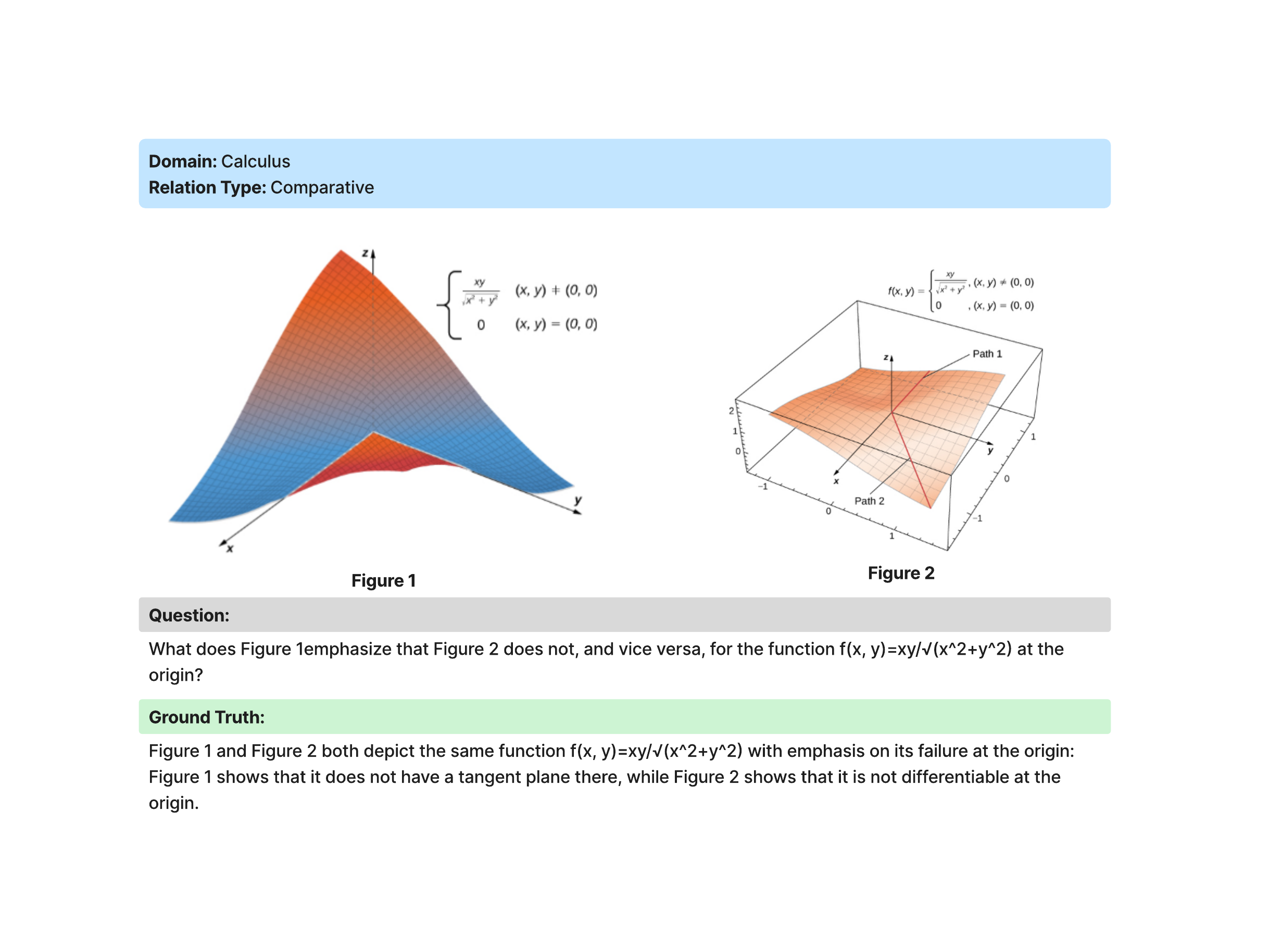}
    \caption{Example of a comparative relational question in the Calculus domain}
    \label{fig:placeholder5}
\end{figure}

\begin{figure}[H]
    \centering
    \includegraphics[width=1\linewidth]{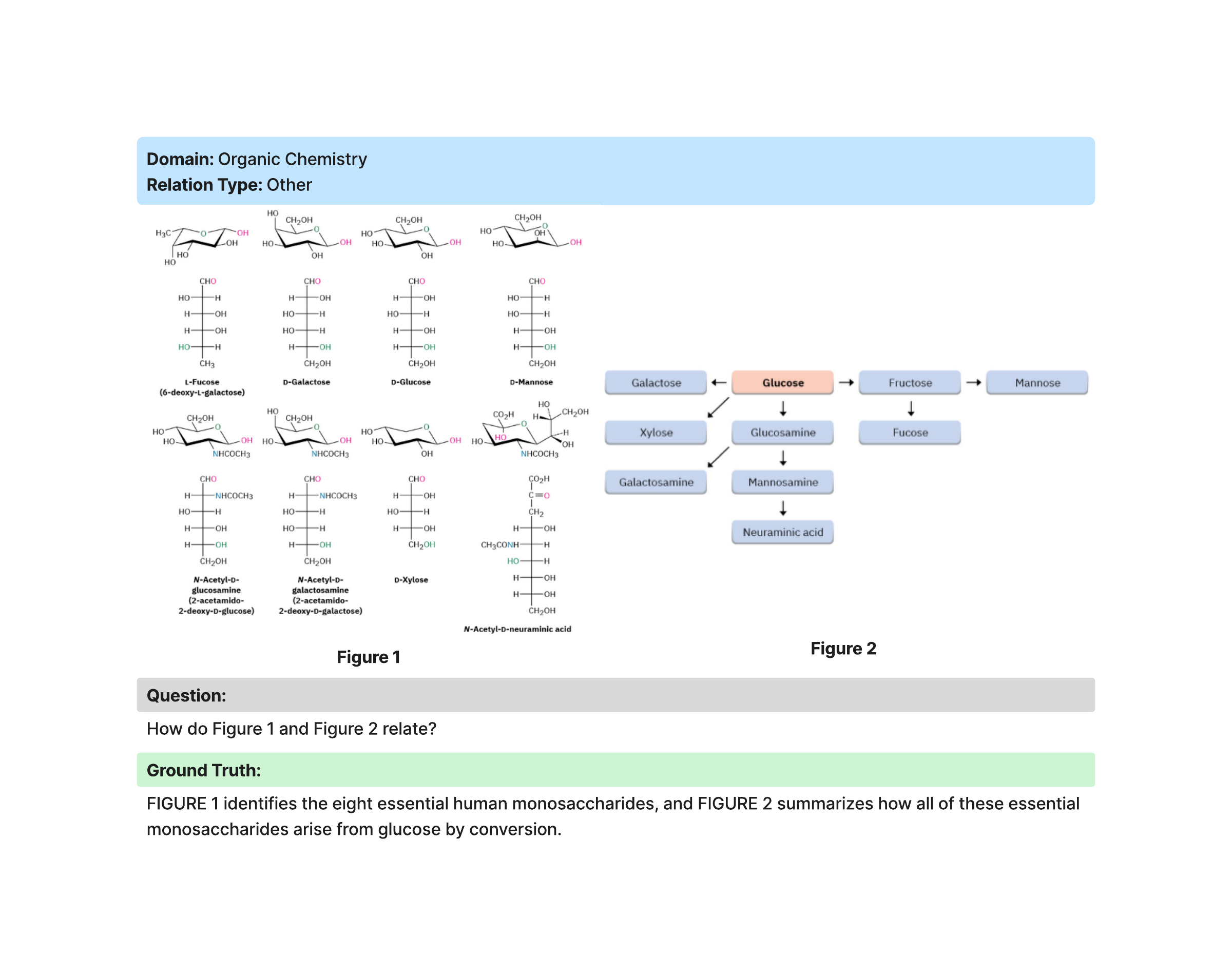}
    \caption{Example of a "Other" type of relational question in the Organic Chemistry domain}
    \label{fig:placeholder6}
\end{figure}

\begin{figure}[H]
    \centering
    \includegraphics[width=1\linewidth]{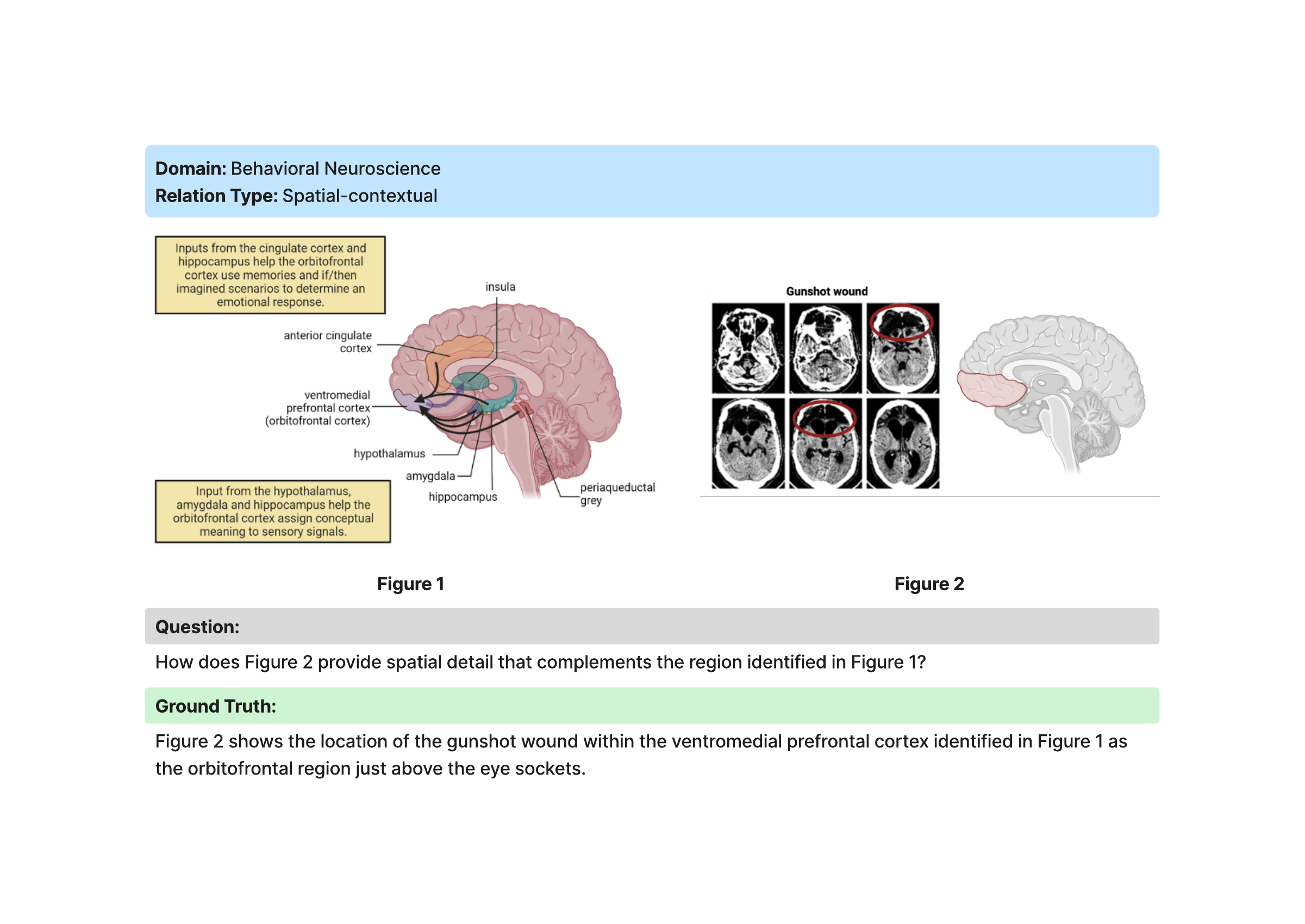}
    \caption{Example of a spatial-contextual relational question in the Behavioral Neuroscience domain}
    \label{fig:placeholder7}
\end{figure}

\begin{figure}[H]
    \centering
    \includegraphics[width=1\linewidth]{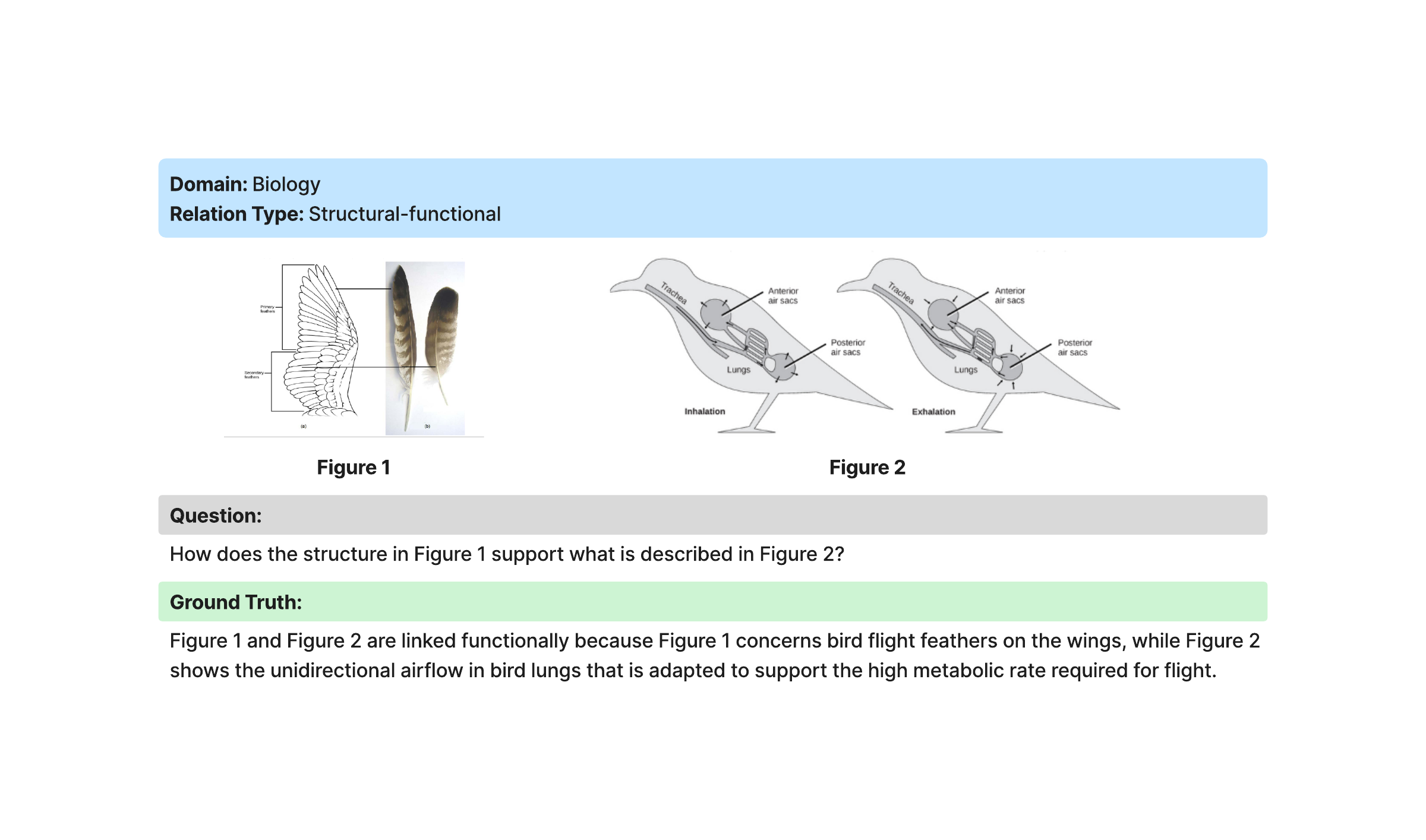}
    \caption{Example of a structural-functional relational question in the Biology domain}
    \label{fig:placeholder8}
\end{figure}

\begin{figure}[H]
    \centering
    \includegraphics[width=1\linewidth]{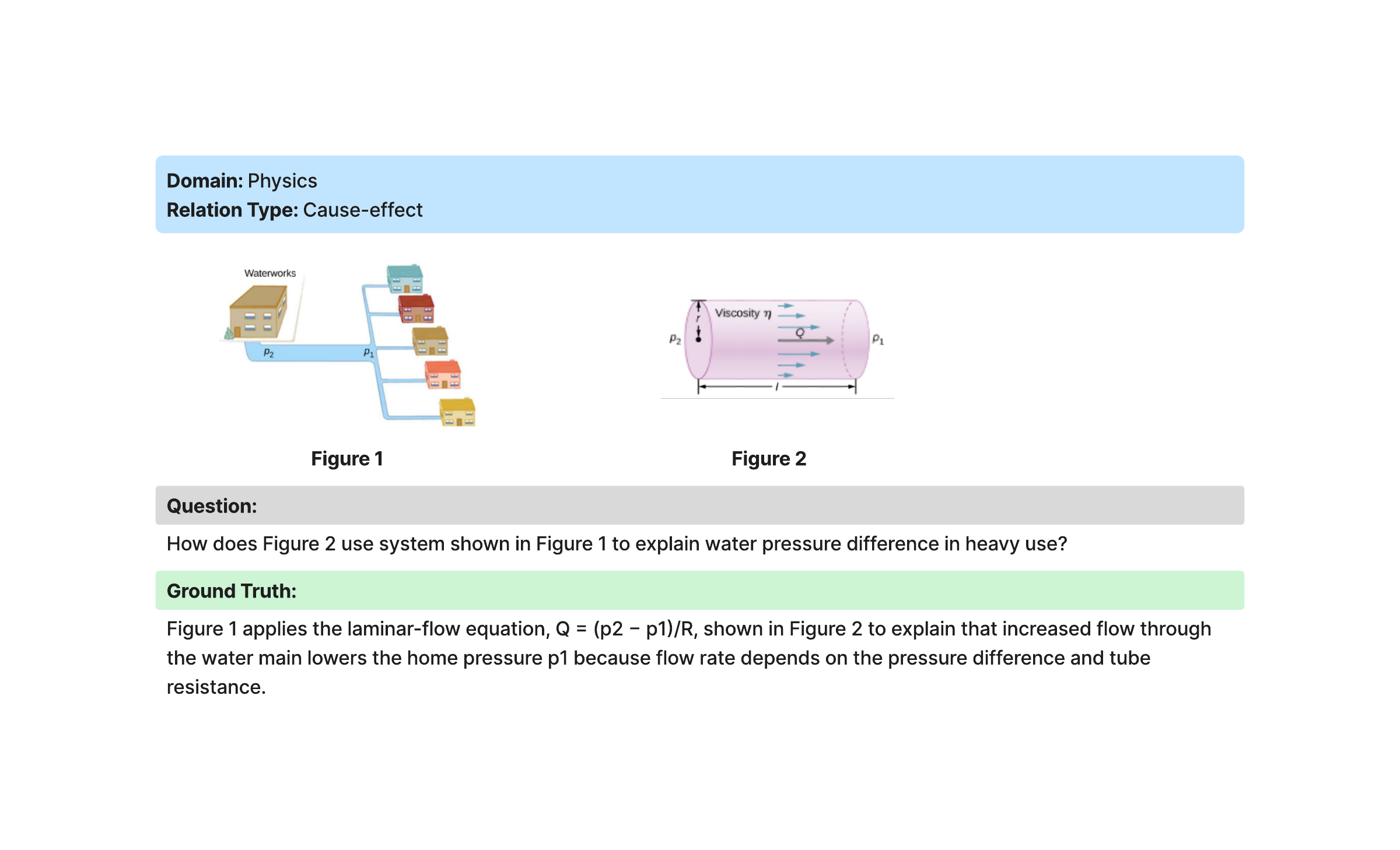}
    \caption{Example of a cause and effect relational question in the Physics domain}
    \label{fig:placeholder9}
\end{figure}

\begin{figure}[H]
    \centering
    \includegraphics[width=1\linewidth]{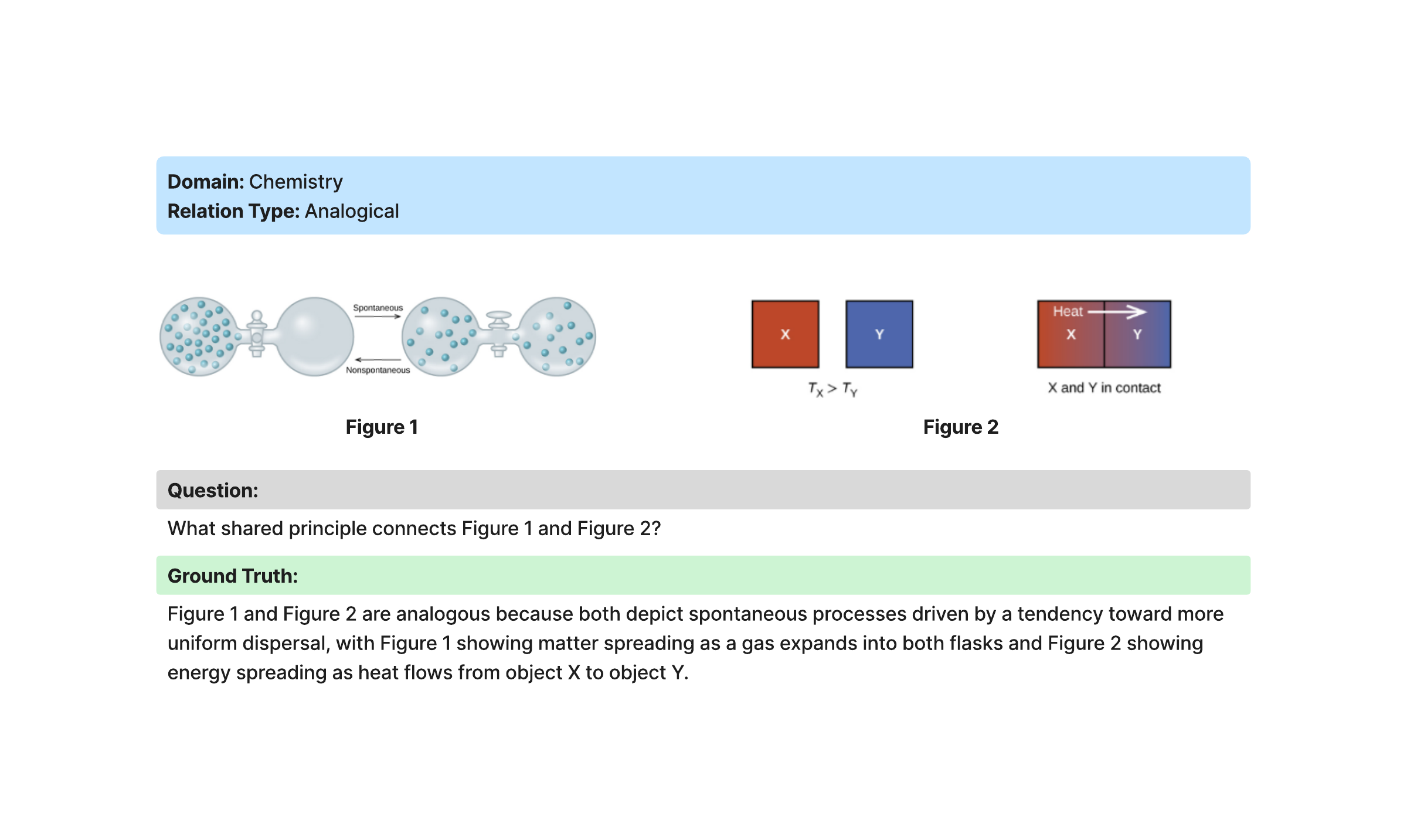}
    \caption{Example of an analogical relational question in the Chemistry domain}
    \label{fig:placeholder10}
\end{figure}

The SciReC varies in terms of relation types and scientific and social subjects. This diversity contributes to a richer contextual representation within the dataset. The combination of relations and knowledge domains provides a comprehensive and diverse evaluation setting. Some examples from the SciReC dataset are shown in Figures \ref{fig:placeholder1}, \ref{fig:placeholder2}, \ref{fig:placeholder3}, \ref{fig:placeholder4}, \ref{fig:placeholder5}, \ref{fig:placeholder6}, \ref{fig:placeholder7}, \ref{fig:placeholder8}, \ref{fig:placeholder9} and \ref{fig:placeholder10}.

\newpage

\section{Prompt and Templates}
\label{prompts}

\begin{figure}[!htbp]
    \centering
    \includegraphics[width=1\linewidth]{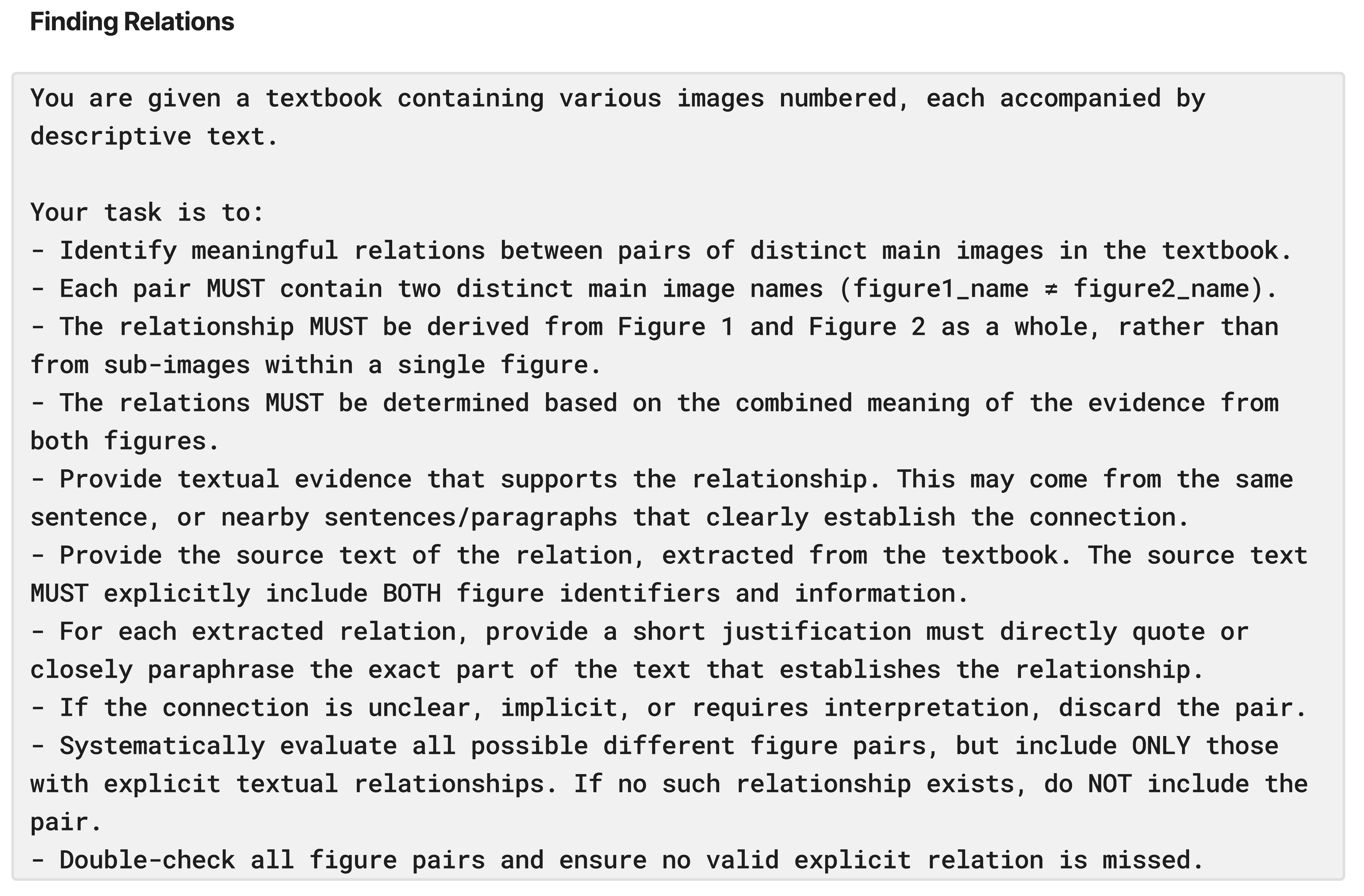}
    \caption{Prompt for extracting the relations from textbooks. }
    \label{fig:pro1}
\end{figure}

\begin{figure}[H]
    \centering
    \includegraphics[width=1\linewidth]{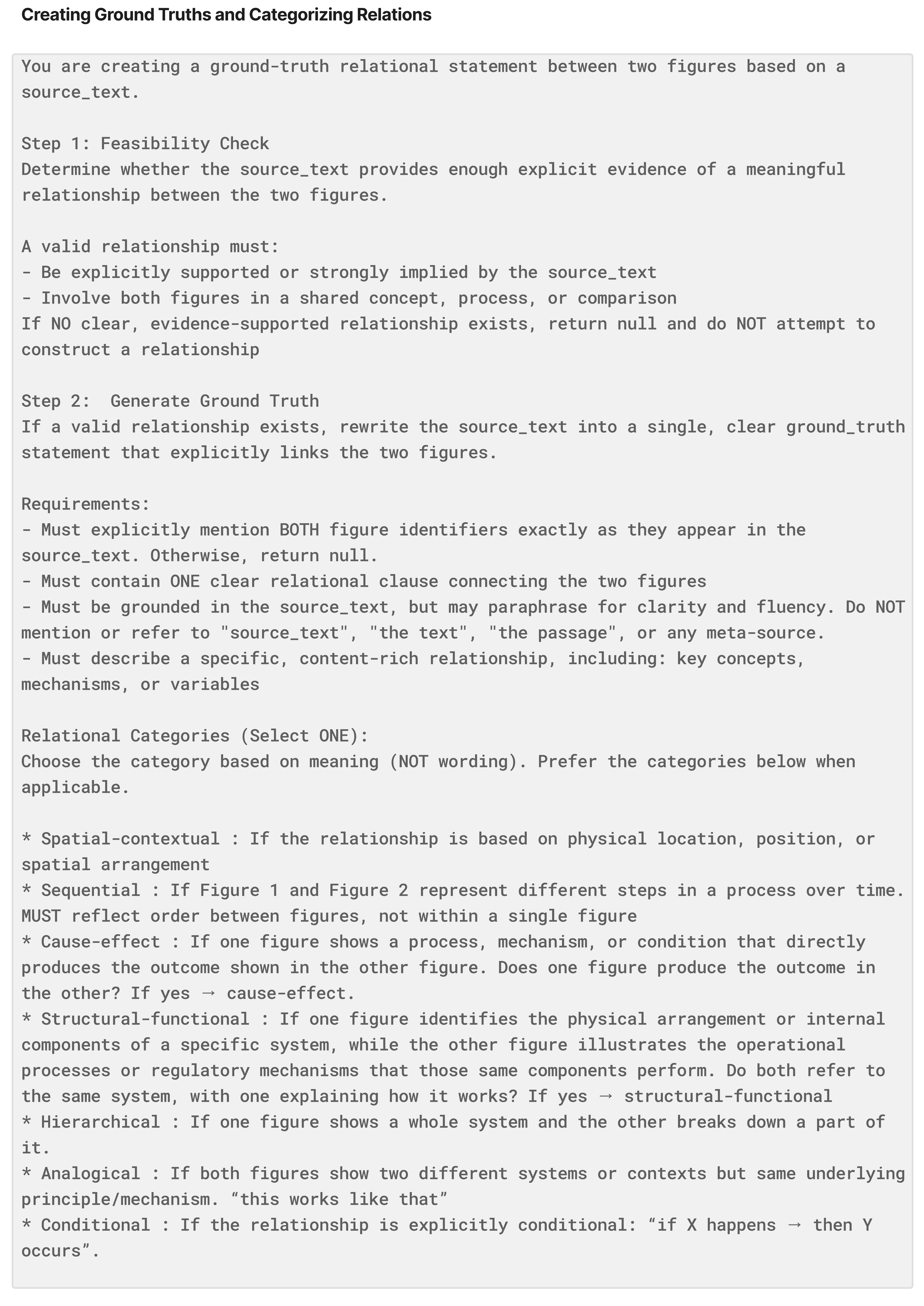}
    \caption{Prompt for relationship validation, ground truth creation, and category selection for relations.}
    \label{fig:pro2}
\end{figure}

\begin{figure}[H]
    \centering
    \includegraphics[width=1\linewidth]{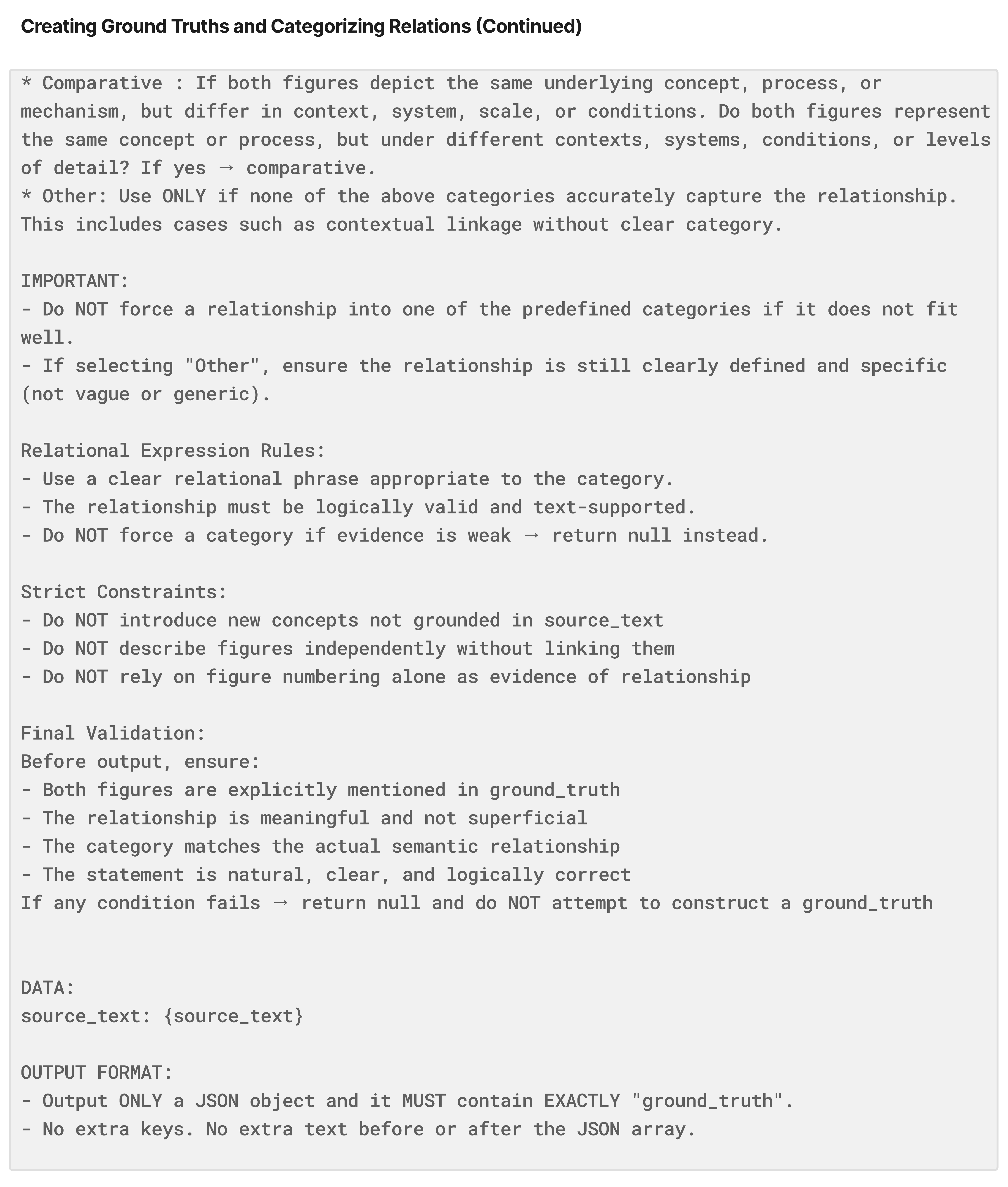}
    \caption{Prompt for relationship validation, ground truth creation, and category selection for relations (continued).}
    \label{fig:pro3}
\end{figure}

\begin{figure}[H]
    \centering
    \includegraphics[width=1\linewidth]{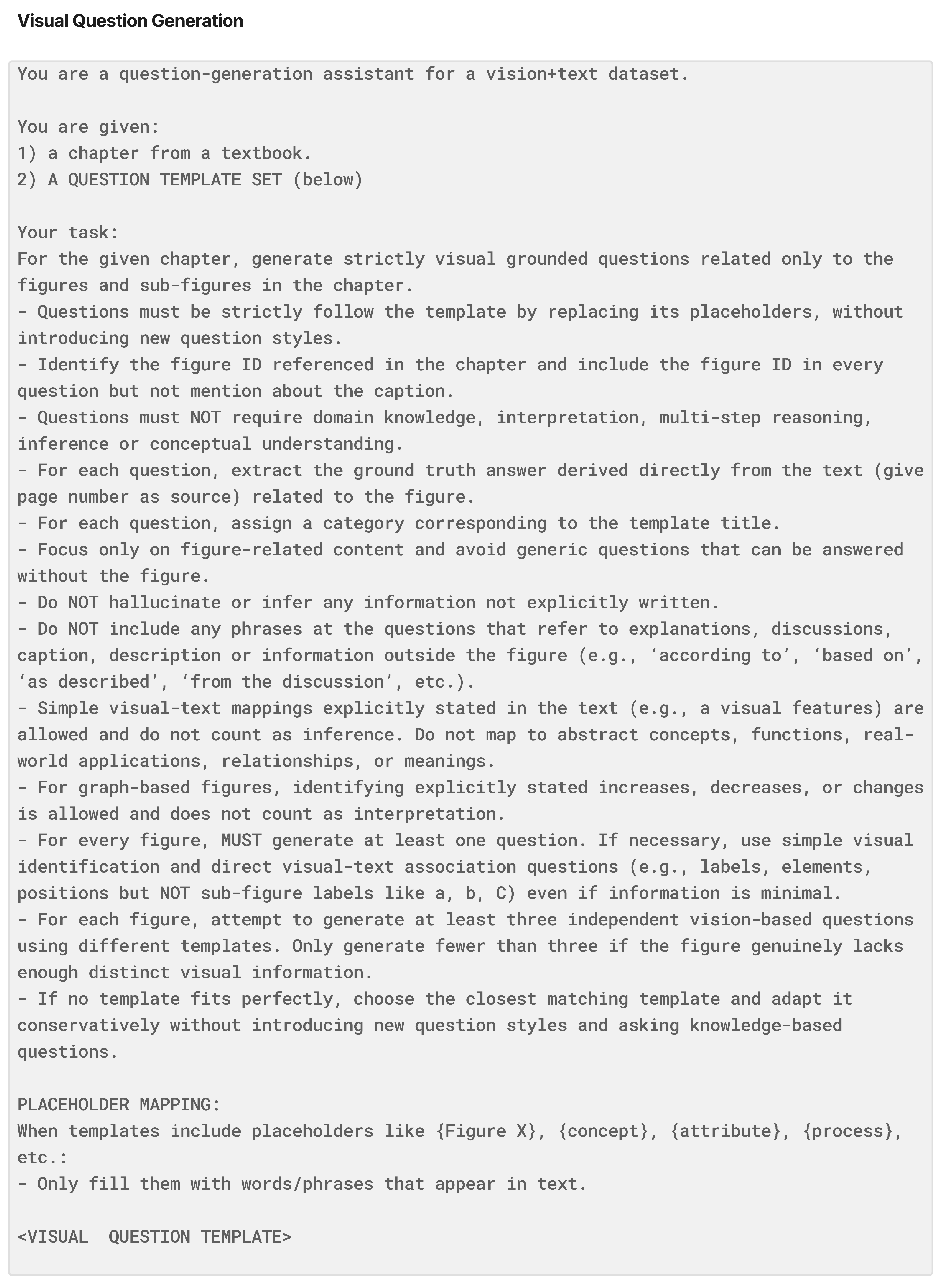}
    \caption{Prompt for creating visual questions with template}
    \label{fig:pro4}
\end{figure}

\begin{figure}[H]
    \centering
    \includegraphics[width=1\linewidth]{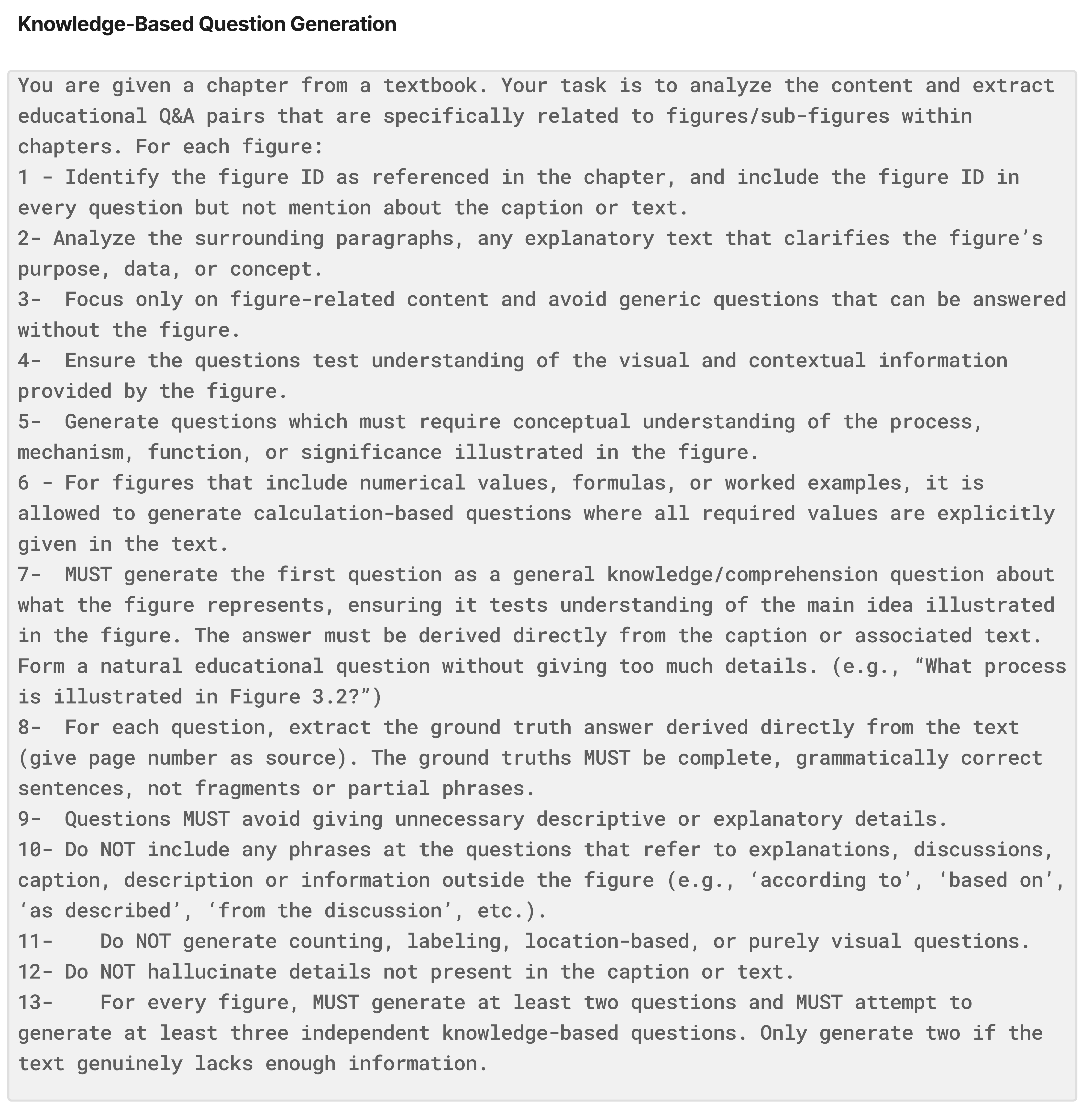}
    \caption{Prompt for creating knowledge-based questions}
    \label{fig:pro5}
\end{figure}

\begin{figure}[H]
    \centering
    \includegraphics[width=1\linewidth]{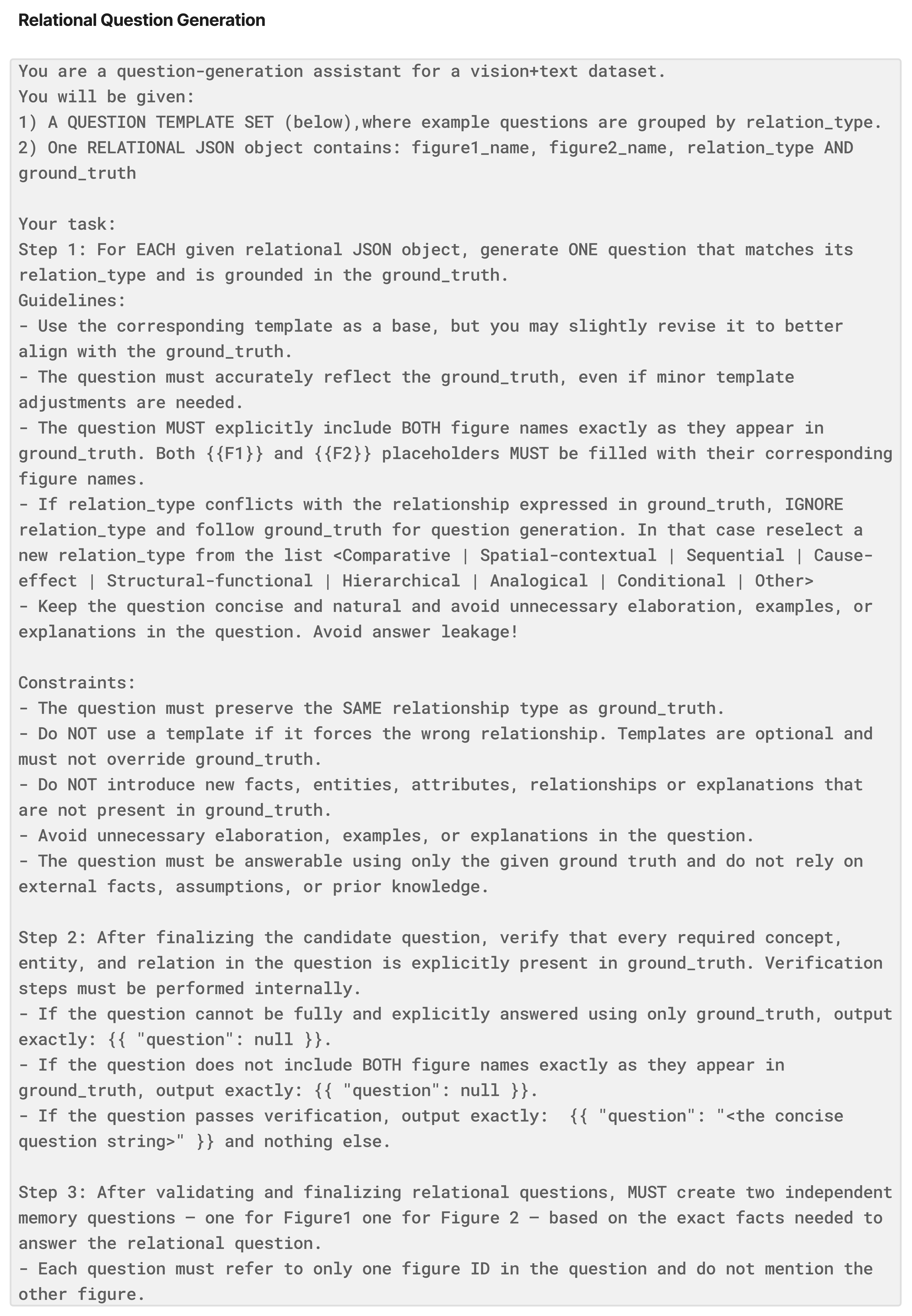}
    \caption{Prompt for creating relational reasoning questions with template.}
    \label{fig:pro6}
\end{figure}

\begin{figure}[H]
    \centering
    \includegraphics[width=1\linewidth]{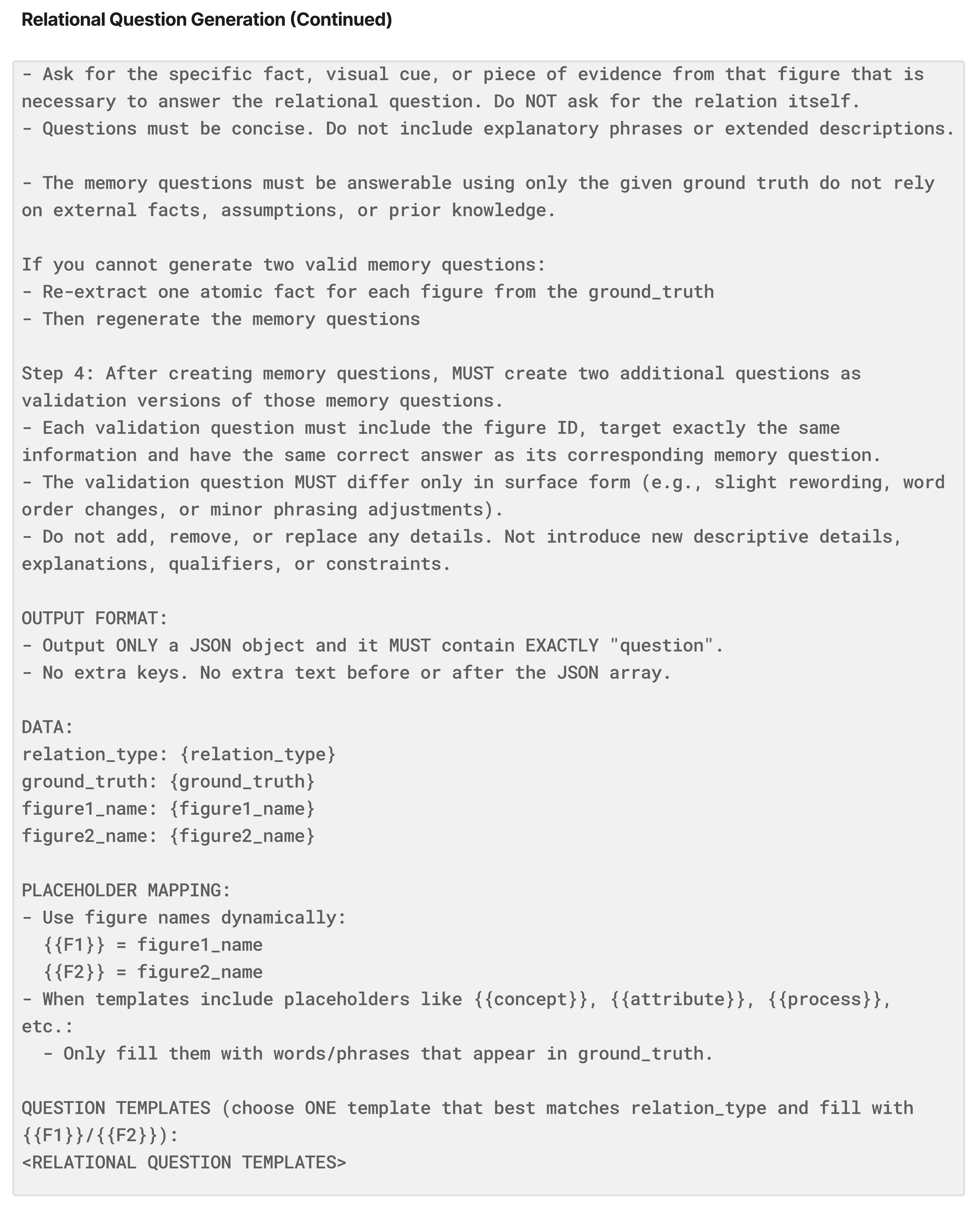}
    \caption{Prompt for creating relational reasoning questions with template (continued).}
    \label{fig:pro7}
\end{figure}

\begin{figure}[H]
    \centering
    \includegraphics[width=1\linewidth]{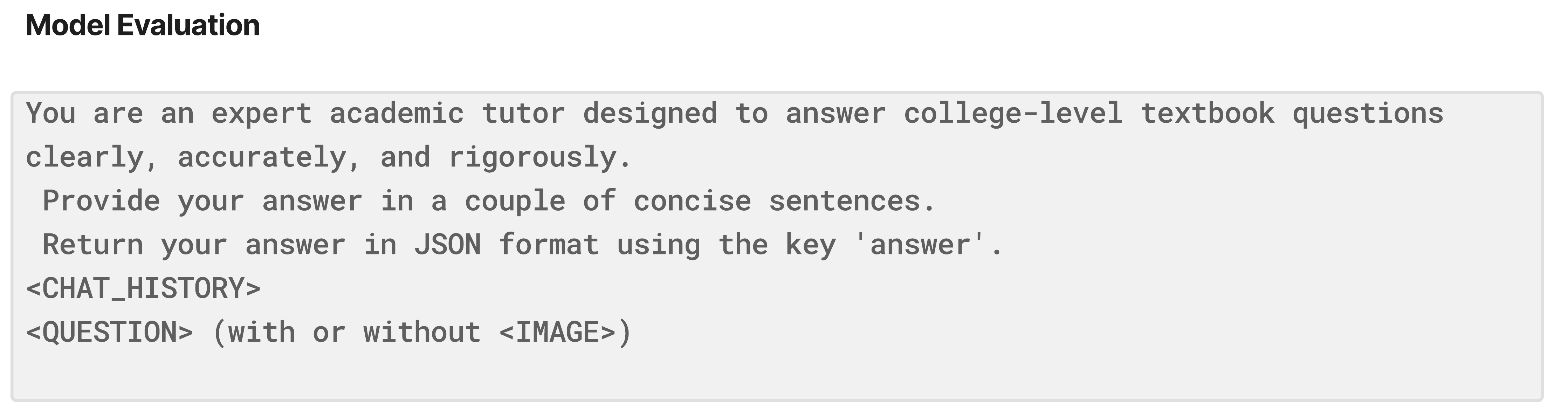}
    \caption{Prompt for model evaluation, including chat history and question-specific images.}
    \label{fig:pro8}
\end{figure}

\begin{figure}[H]
    \centering
    \includegraphics[width=1\linewidth]{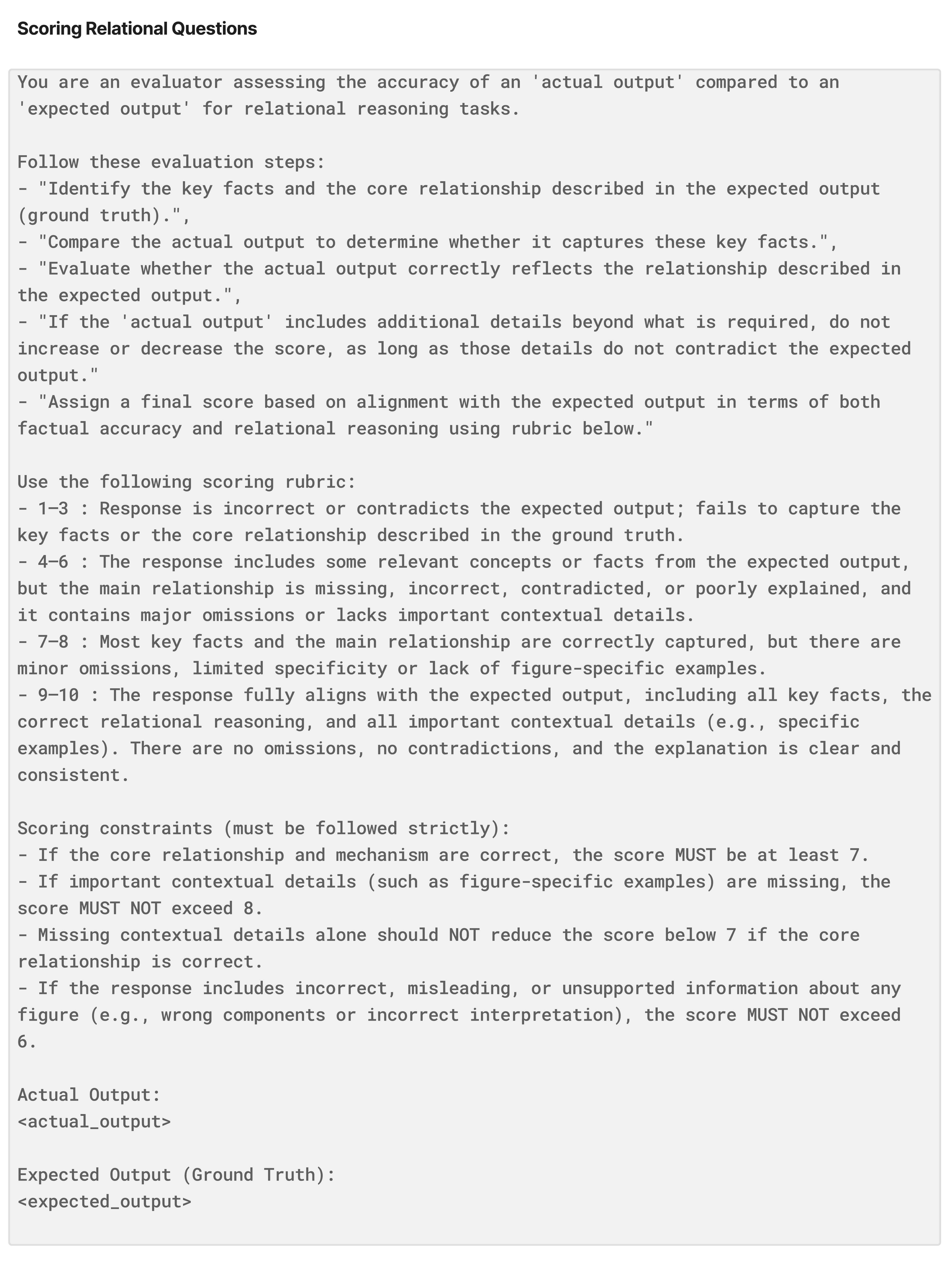}
    \caption{Prompt for scoring model answers in relational reasoning questions by comparing with ground truth using a rubric, scoring from 1 to 10.}
    \label{fig:pro9}
\end{figure}

\begin{figure}[H]
    \centering
    \includegraphics[width=1\linewidth]{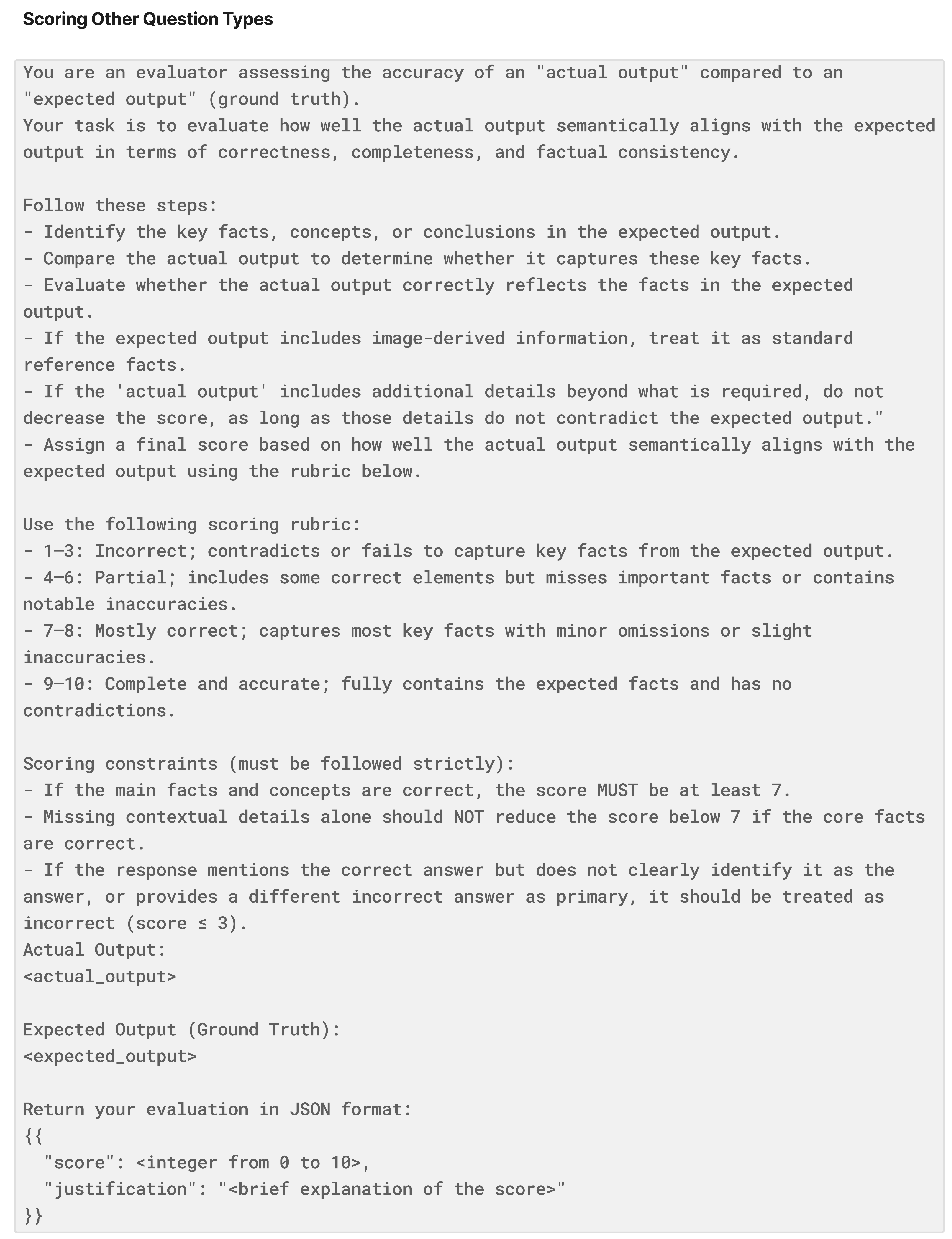}
    \caption{Prompt for scoring model answers in other question types by comparing with ground truth using a rubric, scoring from 1 to 10.}
    \label{fig:pro10}
\end{figure}

\begin{figure}[H]
    \centering
    \includegraphics[width=1\linewidth]{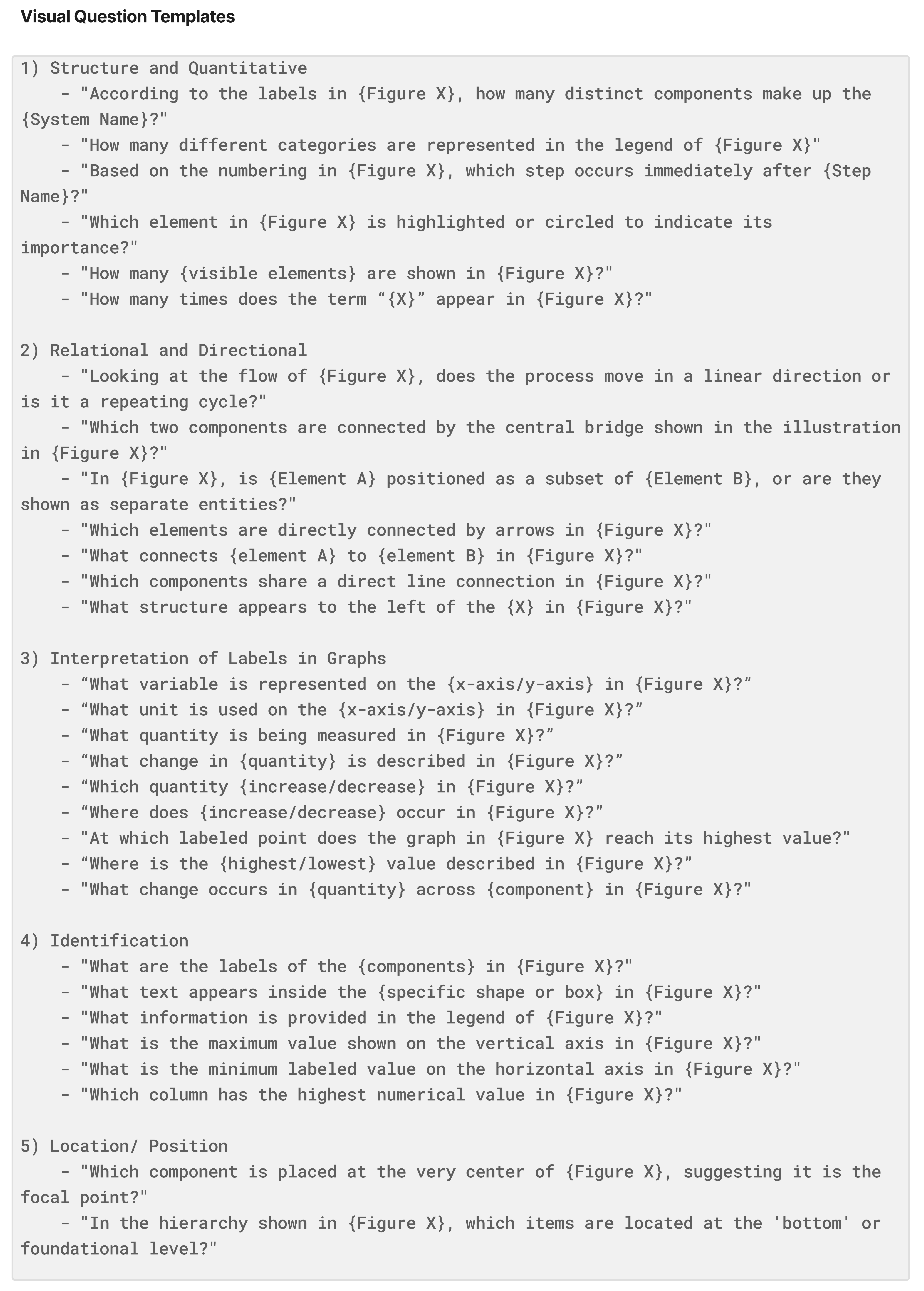}
    \caption{The template for visual questions}
    \label{fig:temp1}
\end{figure}

\begin{figure}[H]
    \centering
    \includegraphics[width=1\linewidth]{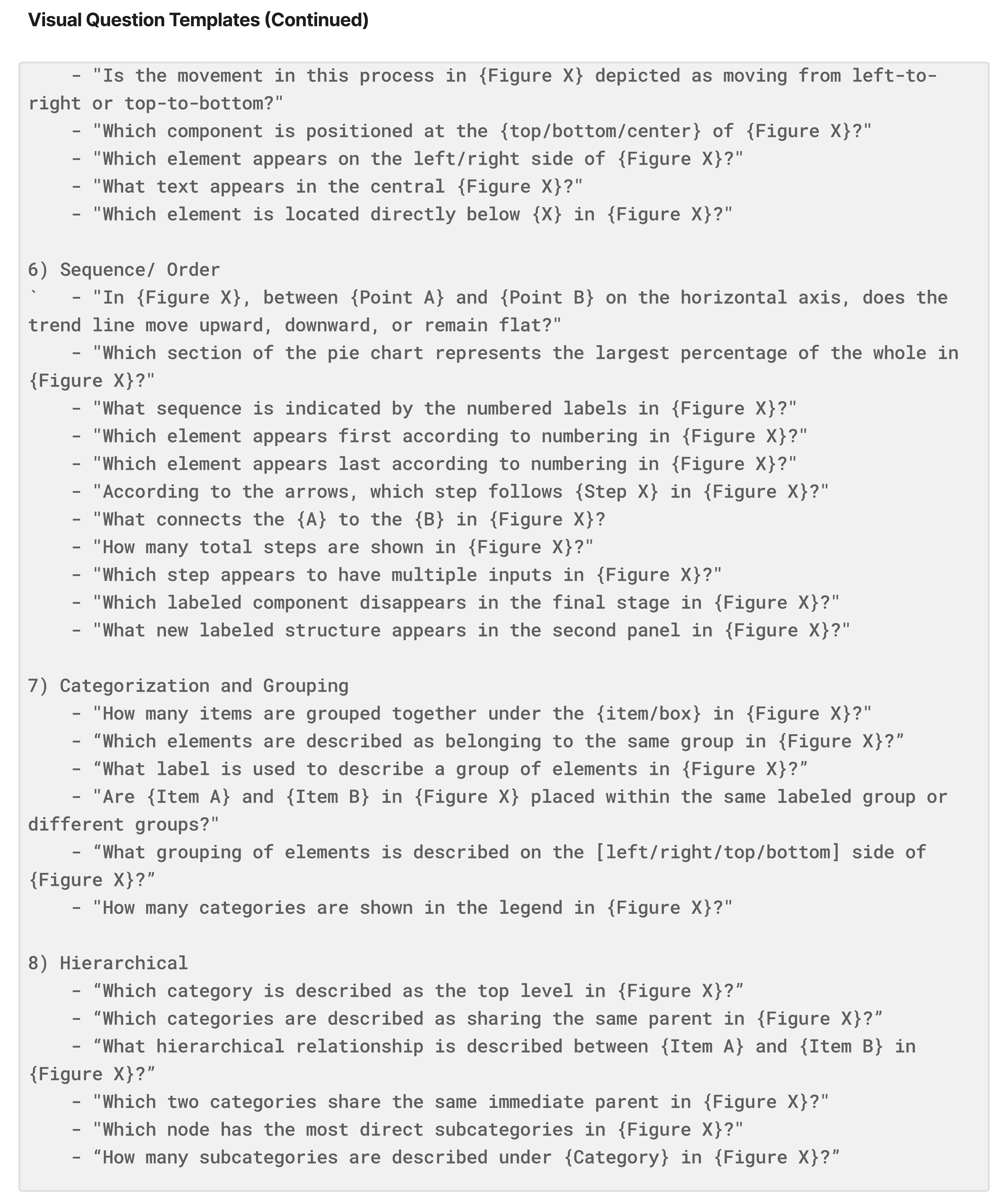}
    \caption{The template for visual questions (continued)}
    \label{fig:temp2}
\end{figure}

\begin{figure}[H]
    \centering
    \includegraphics[width=1\linewidth]{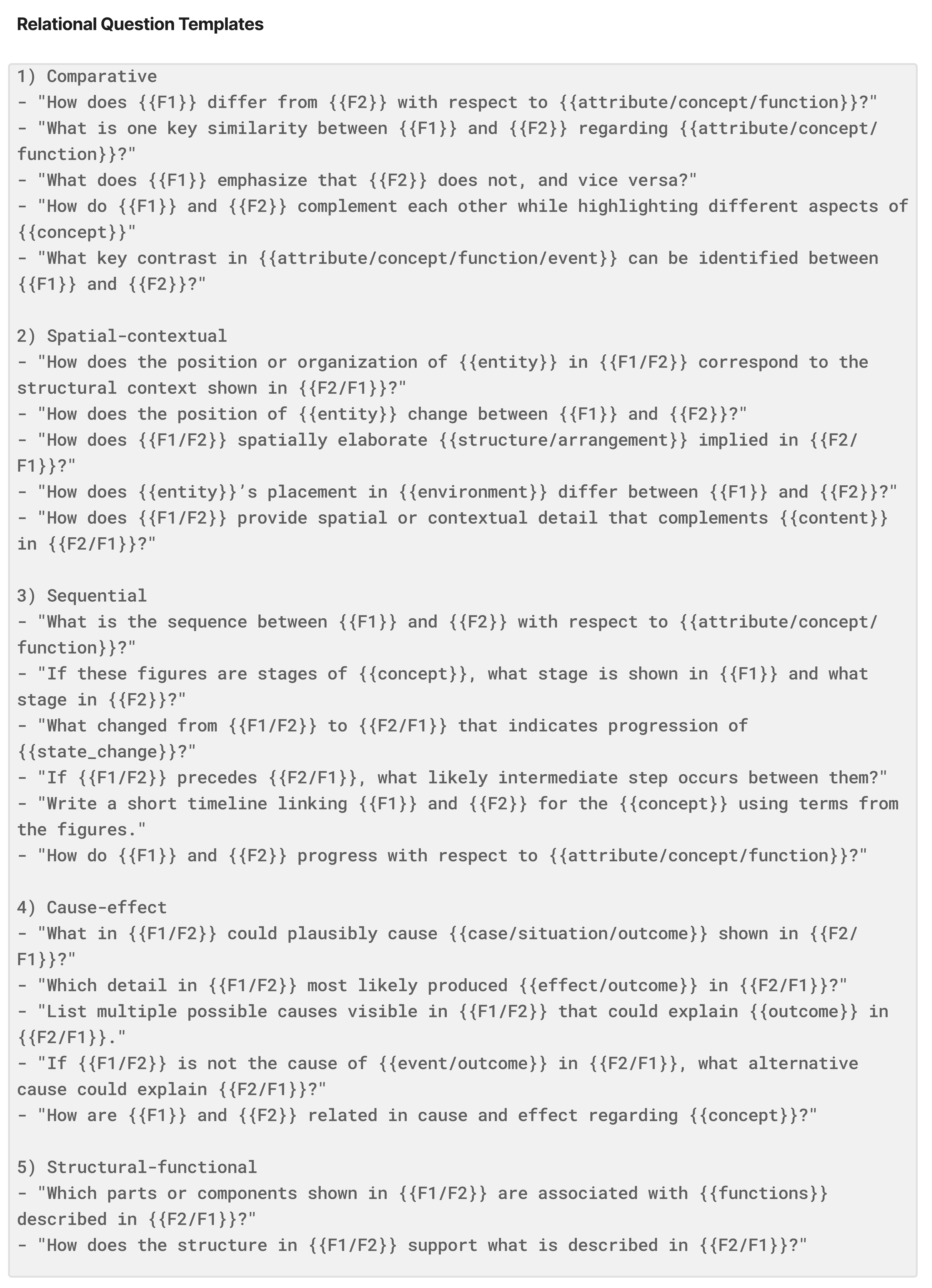}
    \caption{The template for relational questions}
    \label{fig:temp3}
\end{figure}

\begin{figure}[H]
    \centering
    \includegraphics[width=1\linewidth]{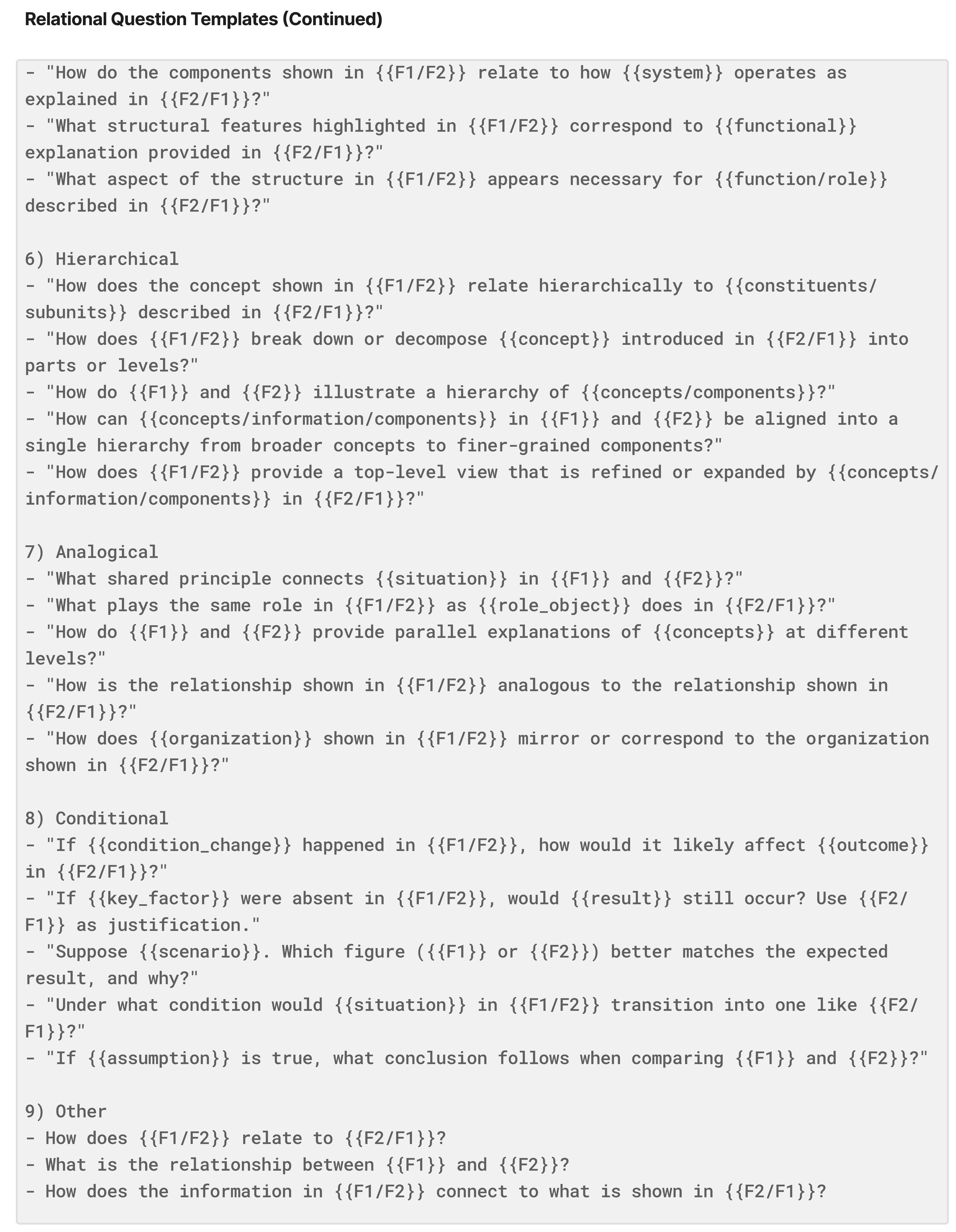}
    \caption{The template for visual questions (continued)}
    \label{fig:temp4}
\end{figure}

The prompts used in the study are provided in Figures \ref{fig:pro1}, \ref{fig:pro2}, \ref{fig:pro3}, \ref{fig:pro4}, \ref{fig:pro5} , \ref{fig:pro6}, \ref{fig:pro7}, \ref{fig:pro8} , \ref{fig:pro9} , \ref{fig:pro10}. The templates utilized to generate visual and relational questions are provided in Figures \ref{fig:temp1}, \ref{fig:temp2}, \ref{fig:temp3}, and \ref{fig:temp4}.

\vspace{20pt}

\clearpage

\subsection{Question Templates}
\label{temp}

\newpage
\section*{NeurIPS Paper Checklist}

\begin{enumerate}

\item {\bf Claims}
    \item[] Question: Do the main claims made in the abstract and introduction accurately reflect the paper's contributions and scope?
    \item[] Answer: \answerYes{} 
    \item[] Justification: Both abstraction and introduction state the paper's contribution along with experiment results.

\item {\bf Limitations}
    \item[] Question: Does the paper discuss the limitations of the work performed by the authors?
    \item[] Answer: \answerYes{} 
    \item[] Justification: The limitation of the work is discussed in Section 6.

\item {\bf Theory assumptions and proofs}
    \item[] Question: For each theoretical result, does the paper provide the full set of assumptions and a complete (and correct) proof?
    \item[] Answer: \answerYes{} 
    \item[] Justification: The theoretical calculations of the DMRA framework are provided in Section 4 along with explanations. A detailed example of the calculations is presented in Appendix C.

    \item {\bf Experimental result reproducibility}
    \item[] Question: Does the paper fully disclose all the information needed to reproduce the main experimental results of the paper to the extent that it affects the main claims and/or conclusions of the paper (regardless of whether the code and data are provided or not)?
    \item[] Answer: \answerYes{} 
    \item[] Justification: The prompt instructions, detailed explanation of the dataset generation pipeline, and calculation of error decomposition are provided in Appendix E, Section 3, and Section 4. Additionally, the code and data are submitted with the paper.

\item {\bf Open access to data and code}
    \item[] Question: Does the paper provide open access to the data and code, with sufficient instructions to faithfully reproduce the main experimental results, as described in supplemental material?
    \item[] Answer: \answerYes{} 
    \item[] Justification: The source link for both data and code is provided in Appendix A.

\item {\bf Experimental setting/details}
    \item[] Question: Does the paper specify all the training and test details (e.g., data splits, hyperparameters, how they were chosen, type of optimizer) necessary to understand the results?
    \item[] Answer: \answerYes{} 
    \item[] Justification: The experiment and test set details are provided in Section 5 and Appendix A. The work does not consist of hyperparameters, optimizers, and training data.

\item {\bf Experiment statistical significance}
    \item[] Question: Does the paper report error bars suitably and correctly defined or other appropriate information about the statistical significance of the experiments?
    \item[] Answer: \answerNo{} 
    \item[] Justification: Error bars are not reported due to the high computational cost.

\item {\bf Experiments compute resources}
    \item[] Question: For each experiment, does the paper provide sufficient information on the computer resources (type of compute workers, memory, time of execution) needed to reproduce the experiments?
    \item[] Answer: \answerYes{} 
    \item[] Justification: The computer resources used in the work are provided in Appendix A.
    
\item {\bf Code of ethics}
    \item[] Question: Does the research conducted in the paper conform, in every respect, with the NeurIPS Code of Ethics \url{https://neurips.cc/public/EthicsGuidelines}?
    \item[] Answer: \answerYes{} 
    \item[] Justification: The research adheres to the NeurIPS Code of Ethics.

\item {\bf Broader impacts}
    \item[] Question: Does the paper discuss both potential positive societal impacts and negative societal impacts of the work performed?
    \item[] Answer: \answerYes{} 
    \item[] Justification: The potential impacts are discussed in Appendix A.
    
\item {\bf Safeguards}
    \item[] Question: Does the paper describe safeguards that have been put in place for responsible release of data or models that have a high risk for misuse (e.g., pre-trained language models, image generators, or scraped datasets)?
    \item[] Answer: \answerYes{} 
    \item[] Justification: The dataset is constructed from publicly available academic sources and does not contain sensitive or personal information. It is released under a  CC BY-NC-SA 4.0 license, which restricts commercial use and requires attribution and share-alike distribution.

\item {\bf Licenses for existing assets}
    \item[] Question: Are the creators or original owners of assets (e.g., code, data, models), used in the paper, properly credited, and are the license and terms of use explicitly mentioned and properly respected?
    \item[] Answer: \answerYes{} 
    \item[] Justification: We have properly credited the textbook writers and publishers used in the study, and adhere to the CC BY-NC-SA 4.0 licensing requirements. 

\item {\bf New assets}
    \item[] Question: Are new assets introduced in the paper well documented and is the documentation provided alongside the assets?
    \item[] Answer: \answerYes{} 
    \item[] Justification: All new assets introduced in the paper are accompanied by clear documentation and usage instructions.

\item {\bf Crowdsourcing and research with human subjects}
    \item[] Question: For crowdsourcing experiments and research with human subjects, does the paper include the full text of instructions given to participants and screenshots, if applicable, as well as details about compensation (if any)? 
    \item[] Answer: \answerNA{} 
    \item[] Justification: This paper does not involve crowdsourcing or human-subject research. Human annotators are used only for internal quality control and dataset verification, not for conducting experiments or collecting human-subject data.

\item {\bf Institutional review board (IRB) approvals or equivalent for research with human subjects}
    \item[] Question: Does the paper describe potential risks incurred by study participants, whether such risks were disclosed to the subjects, and whether Institutional Review Board (IRB) approvals (or an equivalent approval/review based on the requirements of your country or institution) were obtained?
    \item[] Answer: \answerNA{} 
    \item[] Justification: The project does not involve human-subjects research as defined by standard research-ethics rules because the work did not (a) collect data through interaction or intervention with living individuals, nor (b) collect identifiable private information about individuals. The images in benchmarks are extracted from open-source academic textbooks. The only human process involved in dataset construction was data validation/annotation (reviewers validated relational questions). The released benchmark and diagnostic framework do not include personal data or copyrighted images.

\item {\bf Declaration of LLM usage}
    \item[] Question: Does the paper describe the usage of LLMs if it is an important, original, or non-standard component of the core methods in this research? Note that if the LLM is used only for writing, editing, or formatting purposes and does \emph{not} impact the core methodology, scientific rigor, or originality of the research, declaration is not required.
    \item[] Answer: \answerYes{} 
    \item[] Justification: The LLM usage is described in Appendix A.

\end{enumerate}

\end{document}